\documentclass[MS]{macro/neu_msthesis}

\usepackage{physics}
\usepackage{siunitx}
\usepackage{graphics} 
\usepackage{graphicx} 
\usepackage{epsfig} 
\usepackage{lipsum} 
\usepackage{amsmath, bm} 
\usepackage{amssymb}  
\usepackage{upgreek}
\usepackage{url}
\usepackage[breaklinks]{hyperref}
\usepackage{breakurl}

\usepackage{caption}
\usepackage{subcaption}
\usepackage{pgfplots}
\usepackage{floatrow}
\usepackage{float}

\usepackage[style=ieee]{biblatex}

\newfloatcommand{capbtabbox}{table}[][\FBwidth]
\bibliography{thesis}

\title{Feedback Design and Implementation for Integrated Posture Manipulation and Thrust Vectoring}

\author{Aniket Shashikant Dhole}

\dept{Electrical and Computer Engineering}

\degree{Master of Science}

\degreename{Electrical and Computer Engineering}

\field{Electrical and Computer Engineering}

\submitdate{December 2024}

\numberofmembers{3}

\principaladviser{Dr. Alireza Ramezani}
\firstreader{Dr. Miriam Leeser}
\secondreader{Dr. Michael Everett}

\chairman{Dr. Josep M. Jornet}

\dean{Sagar Kamarthi}

\usepackage{amsfonts}			
\usepackage{times}		

\newcommand{\ifno}[1]{}

\usepackage{multirow}
\makeindex
\usepackage{makeidx}
\usepackage{acronym}

\usepackage{url}
\hypersetup{				
pdfauthor = {\authorRef},
pdftitle = {\titleRef},
pdfsubject = {\expandafter{\degreeRef} thesis submitted to Northeastern University},
pdfkeywords = {add keywords here}
pdfcreator = {LaTeX with hyperref package},
pdfproducer = {dvips + ps2pdf}}

\begin{document}

\pdfbookmark[1]{Cover}{cover}

\titlepage
\begin{frontmatter}

\pdfbookmark[1]{Table of Contents}{contents}
\tableofcontents
\listoffigures
\newpage\ssp
\listoftables



\begin{acknowledgements}

This thesis would not have been possible without the invaluable support and guidance of the incredible individuals at the SiliconSynapse Lab at Northeastern University. I am deeply grateful to my advisor, Prof. Alireza Ramezani, whose unwavering belief in my abilities and mentorship from the very start of my graduate journey have been pivotal. His vision, guidance, and encouragement have not only shaped my research but also allowed me to work on challenging and exciting projects that have fueled my passion for robotics. I extend my heartfelt thanks to my lab mates: Bibek, Shreyansh, Eric, Kaushik, Chenghao, Adarsh, Taoran, Xuejian, and Yizhe for their invaluable assistance with hardware development, experiments, control design, and simulations. Your collaboration, insights, and camaraderie have made this journey both productive and enjoyable. I am equally grateful to Prof. Miriam Leeser and Prof. Michael Everett for their support as members of my thesis committee. Your thoughtful feedback and perspectives have been instrumental in refining and elevating my work.
\\
\\
Finally, I owe everything to my family and friends, whose unwavering support, encouragement, and belief in me have been the cornerstone of my achievements. You have stood by me every step of the way, inspiring me to strive for excellence in all that I do. To all of you, thank you for being a part of this journey. This work is as much yours as it is mine.

\end{acknowledgements}


\begin{abstract}
This MS thesis outlines my contributions to the closed loop control and system integration of two robotic platforms: 1) Aerobat, a flapping wing robot stabilized by air jets, and 2) Harpy, a bipedal robot equipped with dual thrusters. Both systems share a common theme of the integration of posture manipulation and thrust vectoring to achieve stability and controlled movement. For Aerobat, I developed the software and control architecture that enabled its first untethered flights. The control system combines flapping wing dynamics with multiple air jet stabilization to maintain roll, pitch and yaw stability. These results were published in the IEEE/RSJ International Conference on Intelligent Robots and Systems (IROS). For Harpy, I implemented a closed-loop control framework that incorporates active thruster assisted frontal dynamics stabilization . My work led to preliminary untethered dynamic walking. This approach demonstrates how thrust assisted stability can enhance locomotion in legged robots which has not been explored before.
\end{abstract}

\begin{center}
    {\fontsize{20}{30}\selectfont\textbf{Thesis Contribution}}
\end{center}

My thesis consists of work I have done on two platforms: Aerobat \cite{dhole_hovering_2023} and Harpy \cite{pitroda_dynamic_2023} at SiliconSynapse Lab, Northeastern University. I mostly worked on the Software Development, Controls and Electronics Prototyping. So my thesis is structured into two chapters regarding these two platforms:

\begin{enumerate}
    \item Software Development of position and attitude control algorithms for Aerobat Guard
    \item Controls and Experimentation of Flight and Gait control for Dynamic Thruster-Assisted trotting on Harpy
    \item Electronics circuit design and sensor integration across both robotic platforms
    \item Robust communication pipelines for high frequency sensor data transmission among multiple subsystems
    \item Integration of ORBSLAM for feature detection and mapping on Aerobat using RPI Camera
\end{enumerate}

\end{frontmatter}


\pagestyle{headings}


\chapter{Aerobat: Morphing Wing Flight Control}
\label{chap:aerobat}

\section{Introduction}
\label{chap:aerobat:intro}

\begin{figure}
\vspace{0.08in}
    \centering
    \includegraphics[width=1\linewidth]{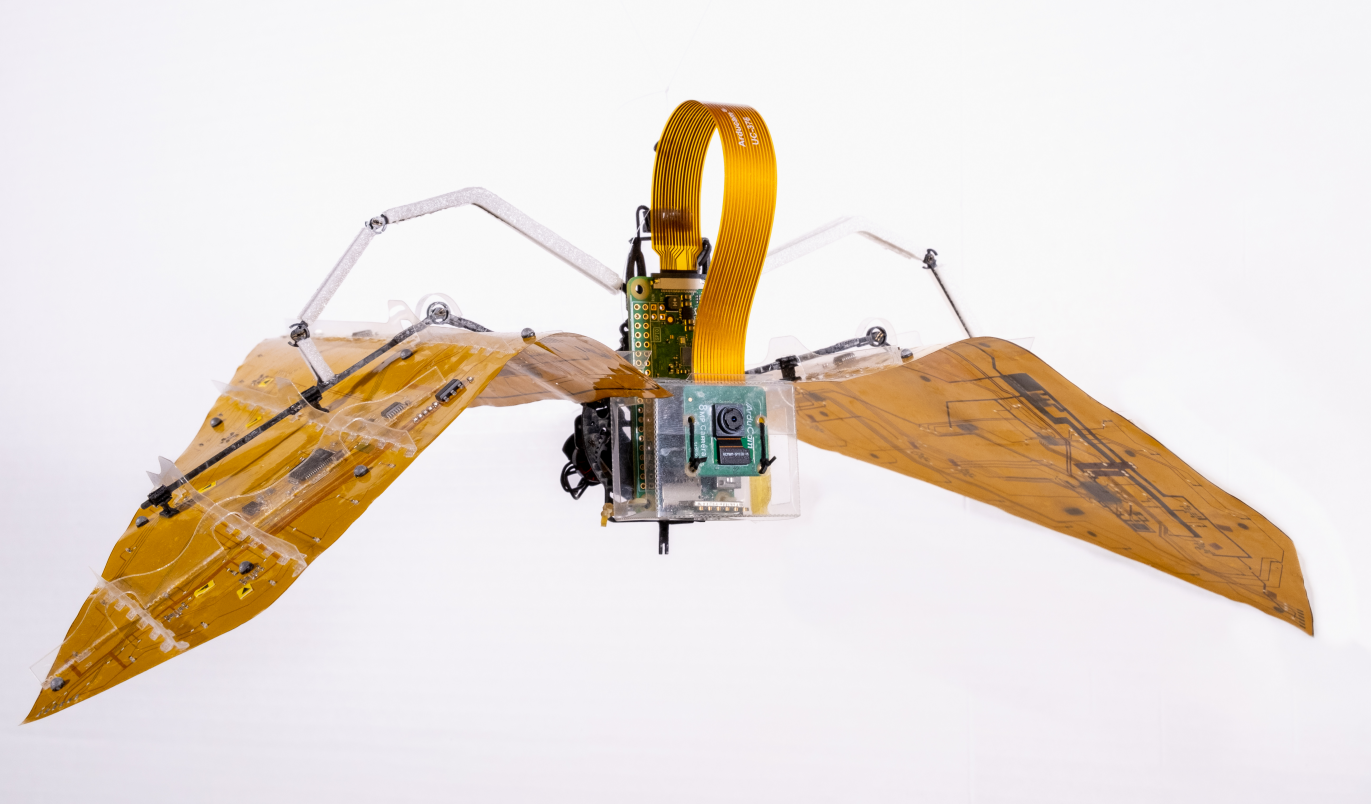}
    \caption{Illustrates Aerobat and its articulated wings that can dynamically reconfigure during a gait cycle.}
    \label{fig:cover}
\vspace{-0.08in}
\end{figure}

This thesis briefly reports my attempts to stabilize the flight dynamics of Northeastern's Aerobat platform \cite{sihite_computational_2020} (Fig.~\ref{fig:cover}). 
This tailless platform can dynamically adjust its wing platform configurations during gaitcycles adding to its efficiency and agility. That said, Aerobat's morphing wings add to the inherent complexity of flight control and stabilization. This work continues my past efforts to control Aerobat's flight dynamics.

Flapping robots are not new. A plethora of examples is mainly dominated by insect-style design paradigms that are passively stable. Passive flight stabilization has a long history in flapping wing flight. Some systems device a tail \cite{de_croon_design_2009, rosen_development_2016, wissa2015free, chang2020soft, send2012artificial}, or design for a low center of mass to achieve open loop stability \cite{phan_kubeetle-s_2019, ma_controlled_2013}. However, tailless flapping flight is relatively new \cite{ramezani_biomimetic_2017}. 
Its advent coincided with the introduction of small sensors and powerful computational resources that these aerial robots could carry for active flight control. 

Aerobat, in addition to being tailless, unlike \cite{chukewad_robofly_2021, de_croon_design_2009, wissa2015free, chang2020soft, send2012artificial, ma_controlled_2013, phan_kubeetle-s_2019, rosen_development_2016}, is capable of significantly morphing its wing structure during each gait cycle. The robot has a weight of 40g when carrying a battery and a basic microcontroller, with an additional payload capacity of 20g, and a wingspan of 30 cm. Aerobat is powered by a 2-cell Lithium Polymer battery and is controlled onboard through a Raspberry Pi Zero 2w that also interacts with its camera and inertial measurement unit (IMU), used for autonomous localization through visual-inertial techniques.

The notion of embodiment in flapping wings, first was introduced and tested experimentally in \cite{ramezani_biomimetic_2017}. 
However, Aerobat possesses more complex embodied reconfiguration capabilities. Aerobat utilizes a computational structure called the \textit{Kinetic Sculpture} (KS) \cite{sihite_computational_2020, hoff_optimizing_2018}, which introduces computational resources for wing morphing. The KS is designed to actuate the robot's wings as it is split into two wing segments: the proximal and distal wings.

The morphing has energy efficiency benefits. The wing folding reduces the wing surface area and minimizes the negative lift during the upstroke, resulting in a more efficient flight. However, wing folding makes Aerobat very unstable. We have attempted outdoor flight using launchers to demonstrate Aerobat's thrust generation capabilities at high speeds \cite{siliconsynapse_lab_progress_2022}. 
However, maintaining a fixed position and orientation in hovering modes is ideally desirable because of the space constraints and diverse application of hovering aerial vehicles.

In this work, I aim to achieve stable hovering maneuvers. Since the actuation framework considered for Aerobat is being designed currently \cite{ramezani_aerobat_2022, sihite2021integrated, sihite_enforcing_2020, ramezani2020towards}, I am attempting other closed-loop operation options based on the addition of external position and orientation compensators to Aerobat using a guard as well. First, I cover the guard design. Next, I explain the model of Aerobat-guard considered in this paper. After, I describe my control approach, followed by results and concluding remarks. 

\begin{figure*}[h!]
\vspace{0.08in}
    \centering
    \includegraphics[width=1\linewidth]{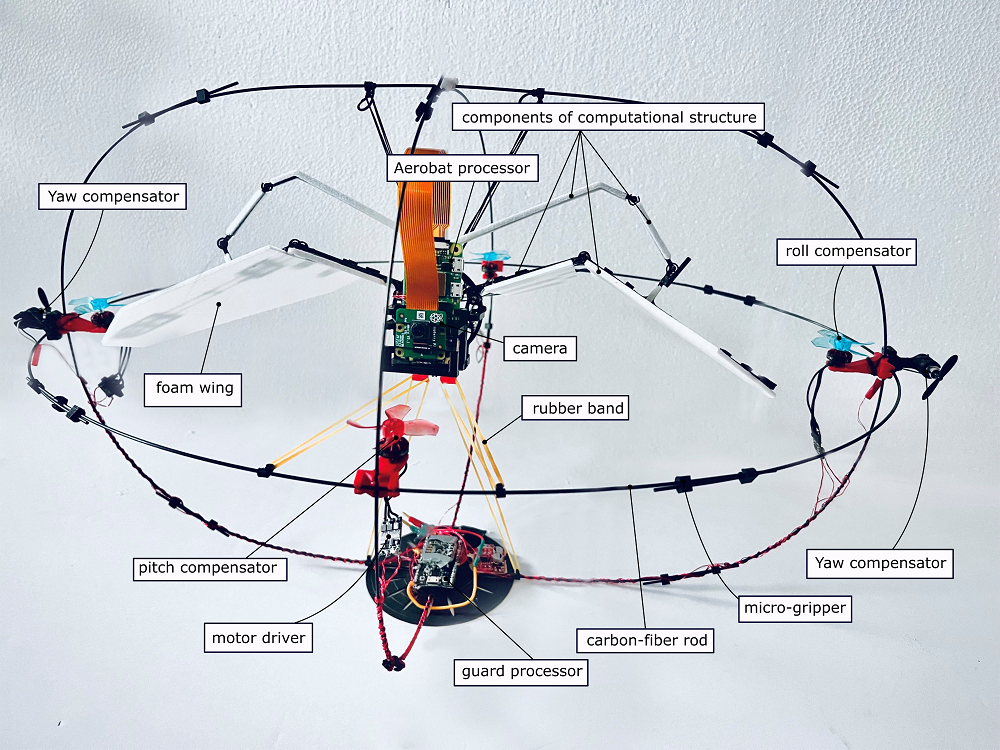}
    \caption{Illustrates Aerobat encapsulated inside a protecting guard.}
    \label{fig:guard}
\vspace{-0.08in}
\end{figure*}

\section{Guard Hardware}

To address the hovering challenge and prepare for the development of closed-loop control, I have built an active stabilization and passive protection system referred to here simply as \textit{The Guard} shown in Fig. \ref{fig:guard}. 

The guard is made up of a set of 11 lightweight 1.5mm carbon fiber rods, attached by PLA-3D printed pieces, and Aerobat is suspended at the center of the guard through four elastic rubber bands. The deformable elastic structure is designed to provide Aerobat with all-around protection in the event of a crash or collision with the environment. To actively stabilize Aerobat, the guard is also equipped with six small DC motors that stabilize the roll, pitch, yaw, and x-y-z positions of the robot. 

These thrusters can carry their own weight, thus nullifying the effect of the added weight of the guard and its electronics; however, they are not powerful to independently lift the system, that is, the lift force generated by Aerobat is required for hovering. To stabilize the yaw dynamics, the guard is equipped with two small and lightweight motors located at the ends of the long axis of the guard (see Fig.~\ref{fig:guard}). Due to the long arms, these motors need not be very powerful and can stabilize the roll, pitch, and yaw angles while adding minimal weight. 

The active stabilization offered by the guard allows Aerobat to test its thrust generation capabilities and, in small increments, exert control of the position and heading of the combined system. Over time, as Aerobat's controls become more stable and reliable through tuning and developing more accurate models, I will reduce the effort the guard applies on active stabilization and let Aerobat take full control of its flight. 

\begin{figure*}[h]
\vspace{0.08in}
    \centering
    \includegraphics[width=1\linewidth]{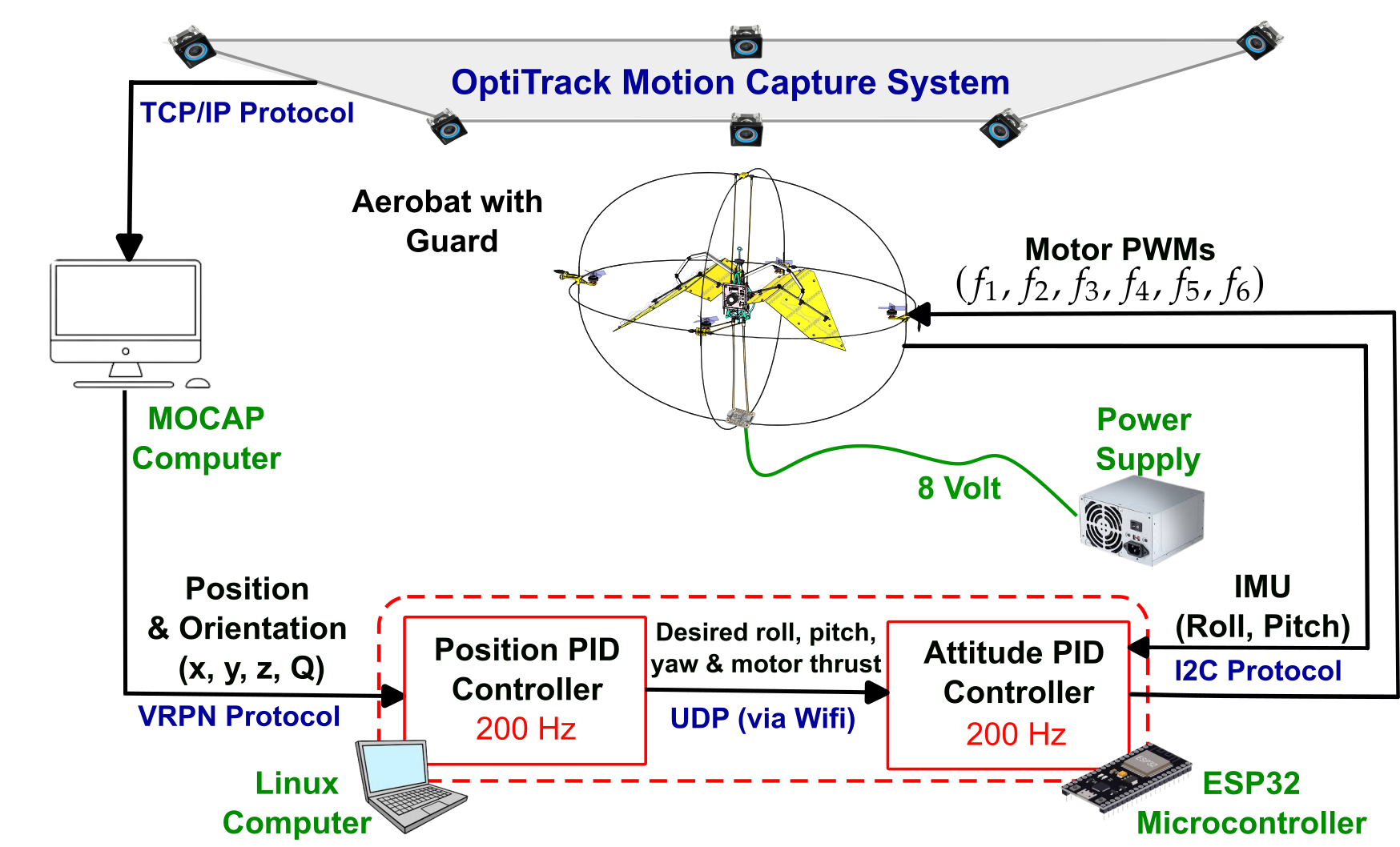}
    \caption{Illustrates Aerobat's Guard Architecture}
    \label{fig:aerobat_arch}
\vspace{-0.08in}
\end{figure*}

The guard is equipped with motion capture markers for closed-loop control and communicates wirelessly with an Optitrack motion capture system. It possesses an onboard micro-controller, ESP32. The ESP32 serves as the primary flight controller for the guard. It features a dual-core Tensilica LX6 microprocessor with clock speeds up to 240 MHz and integrated WiFi connectivity, which is used to transmit data from OptiTrack. The ESP32 controls the six electronic speed controllers. An IMU, ICM-20948, is utilized to estimate the guard's orientation in addition to the OptiTrack position and orientation data.

\subsection{Communication Pipeline}

\textbf{OptiTrack to MoCap Computer Protocol:}
\\
The OptiTrack system operates through a sophisticated network of cameras operating at $240\text{Hz}$, utilizing Power over Ethernet (PoE) for both power delivery and data transmission. Each camera requires $30\text{W}$ (PoE+) power, delivered through Cat6 Ethernet cables. The cameras emit IR light at $850\text{nm}$ wavelength and capture reflections from retroreflective markers with sub-millimeter accuracy ($\pm 0.1\text{mm}$)[3]. The system employs TCP/IP protocol for reliable data transmission between cameras and the MoCap computer, with each camera streaming 2D marker positions at $1\text{Gbps}$ bandwidth. The Motive software processes this multi-camera data with a system latency of approximately $2.8\text{ms}$, maintaining frame synchronization within $\pm 100\text {us}$ across all cameras.
\\
\textbf{VRPN Protocol Implementation}
\\
The Virtual-Reality Peripheral Network (VRPN) protocol facilitates data streaming between the MoCap computer and Linux system at $240\text{Hz}$. VRPN offers two connection modes:
\begin{itemize}
\item \textbf{UDP Mode}: Preferred for low-latency applications, operating on port 3883
\item \textbf{TCP Mode}: Ensures reliable delivery but with higher latency
\end{itemize}
I use the UDP mode as I need more faster data, it transmits 6-DoF pose data including:
$ P = \{x, y, z, q_w, q_x, q_y, q_z\} $
where $(x,y,z)$ represents position and $(q_w,q_x,q_y,q_z)$ represents quaternion orientation.
\\
\textbf{Linux to ESP32 Communication:}
\\
The Linux computer processes position data and transmits control commands to the ESP32 at $200\text{Hz}$ via UDP over WiFi. The control data structure includes:
$ C = \{\theta_{\text{roll}}, \theta_{\text{pitch}}, \theta_{\text{yaw}}, [f_1,...,f_6]\} $
where $\theta$ represents angular commands and $f_i$ represents individual motor thrust values.
\\
\textbf{ESP32 Control Architecture}
\\
The ESP32 implements a dual-control system:
\begin{itemize}
    \item  \textbf{IMU Interface}: Communicates via I2C protocol at $100\text{kHz}$ bus speed, providing roll and pitch measurements with $\pm 0.1°$ accuracy.
    \item  \textbf{Motor Control}: Generates six PWM signals at $50\text{Hz}$ with $16$-bit resolution for precise motor control. The system operates from an $8\text{V}$ power supply, with each motor controller receiving individual thrust commands through dedicated PWM channels.
\end{itemize}
The attitude controller runs at $200\text{Hz}$, processing both IMU data and reference commands from the Linux computer to maintain stable flight dynamics.
\section{Modeling}

I consider the dynamics of the guard and Aerobat as a multi-agent platform that interact. I assume the guard does not know Aerobat's states. 

\begin{figure*}[h!]
\vspace{0.08in}
    \centering
    \includegraphics[width=0.8\linewidth]{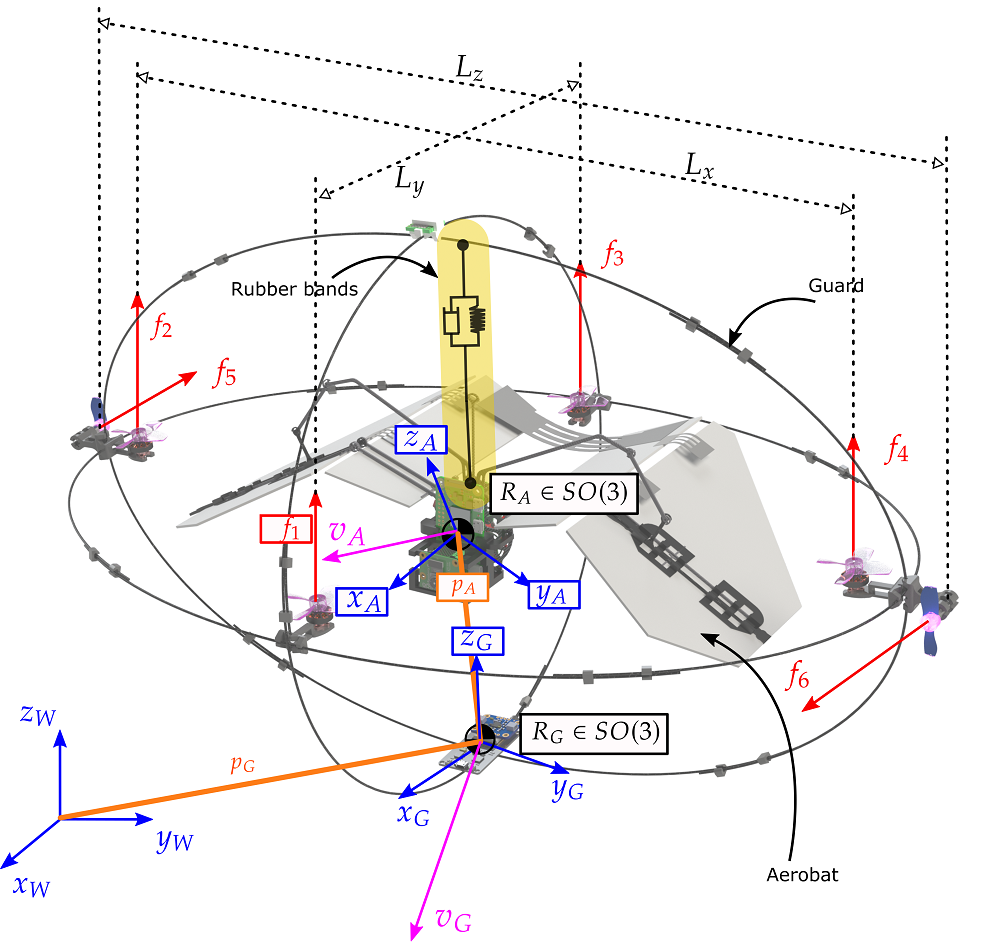}
    \caption{Illustrates Aerobat-guard free-body-diagram.}
    \label{fig:guard-fbd}
\vspace{-0.08in}
\end{figure*}

\begin{figure}[h!]
\vspace{0.08in}
    \centering
    \includegraphics[width=0.4\linewidth]{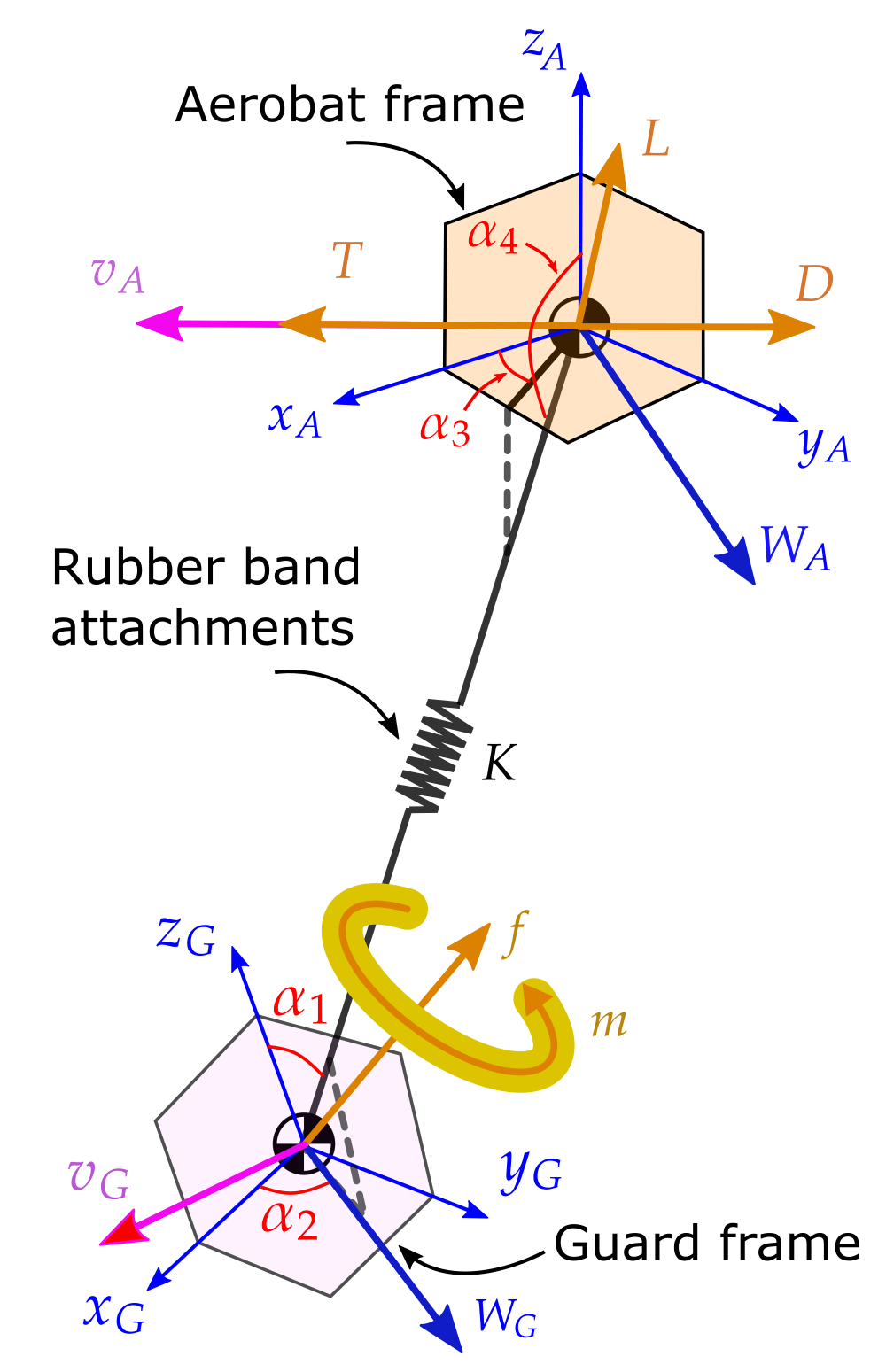}
    \caption{Shows the reduced-order-model (ROM) considered to describe Aerobat's bounding flight. $T$, $L$, and $D$ denote the thrust, lift, and drag forces that are not observable to the guard. Instead, they are added as an extended state ($x_3$ in Eq.~\ref{eq:observer}) to the guard dynamics and its estimated value used to control guard-Aerobat platform.}
    \label{fig:roms}
\vspace{-0.08in}
\end{figure}

I use a reduced-order model (ROM) (see Fig.~\ref{fig:roms}) of the guard-Aerobat platform for control design. The ROM has a total 14 DoFs, including 6 DoFs position and orientations of the guard (fully observable) and a total of 8 DoFs in Aerobat, including the position (3 DoFs), orientations (3 DoFs) relative to the guard, and wing joint angles (2 DoFs due to the symmetry in wings). This model is described with $q$, the configuration variable vector. 

Consider the guard's position $p_G$, orientation (Euler angles) $q_G=[q_x,q_y,q_z]^\top$, body-frame angular velocity vector $\omega_G$, mass $m_G$, and inertia $J_G$. The body-frame forces $f_i$ and moments $m_i$ acting on the guard are:
\begin{equation}
\begin{aligned}
& f=f_1+f_2+f_3+f_4+f_5+f_6+f_e \\
& m_x=L_x\left(f_4-f_2\right)+m_{e,x} \\
& m_y=L_y\left(f_3-f_1\right)+m_{e,y} \\
& m_z=L_z\left(f_6-f_5\right)+m_{e,z}
\end{aligned}
    \label{eq:body-force-moment}
\end{equation}
\noindent where $f_i,~~i=1-6$ are roll, pitch, and yaw compensators as shown in Fig.~\ref{fig:guard-fbd}. $L_i$ is shown in Fig.~\ref{fig:guard-fbd}. In Eq.~\ref{eq:body-force-moment}, $f_e$ denotes the rubber bands' pretension forces. In Eq.~\ref{eq:body-force-moment}, I assume the lift, drag, and thrust forces introduced by the Aerobat have impinged on the guard through the rubber bands' tension forces. While this assumption can be incorrect due to the complex wake interactions between Aerobat's wings and the guard propellers, the assumption permits the isolated modeling of these two systems using standard tools as follows. 

The body forces are mapped to the inertial-frame force by
\begin{equation}
\left[\begin{array}{c}
F_x \\
F_y \\
F_z
\end{array}\right]=R_G^0(q_G)\left[\begin{array}{c}
0 \\
0 \\
f
\end{array}\right]
\end{equation}
\noindent This force is described in the world frame using $R_G^0$. The general equations of motion of the guard while interacting with Aerobat are given:
\begin{equation}
\Sigma_{Guard}:\left\{\begin{array}{l}
\ddot p_G= -g[0,0,1]^\top+\frac{1}{m_G}F\\
\dot R_G^0= R_G^0 \hat\omega_G \\
J_G \dot\omega_G+\omega_G \times J_G \omega_G =m 
\end{array}\right.
    \label{eq:guard-modell}
\end{equation}
\noindent where $g=9.8~ms^{-2}$. 
The equations of motion of Aerobat are given by
\begin{equation}
\Sigma_{Aerobat}:\left\{\begin{array}{l}
[\ddot p_A^\top,\ddot q_A^\top]^\top = -D_u^{-1}\Big( D_{ua} \dot a(t)-H_u+J^\top y\Big)\vspace{0.1in}\\
\dot \xi=\Pi_1(\xi)\xi+\Pi_2(\xi)a(t) \vspace{0.1in}\\
y=\Pi_3(\xi)\xi+\Pi_4(\xi)a(t)
\end{array}\right.
    \label{eq:aerobat-full-dynamics}
\end{equation}
\noindent where $\Pi_i$ and $\xi$ are the aerodynamic model parameters and fluid state vector. $y$ is the output of the aerodynamic model, that is, the aerodynamic force. $J$ is the Jacobian matrix. $a(t)=[\dots a_i(t)\dots]^\top$ denotes wings' joint trajectories. $p_A$ and $q_A$ denote the position and orientation of Aerobat with respect to the guard and are given by
\begin{equation}
    \begin{aligned}
        p_A = R_A^G(q_A)[0,0,1]^\top\\
        q_A = [\alpha_3,\alpha_4]^\top
    \end{aligned}
\end{equation}
\noindent Please see Fig.~\ref{fig:guard-fbd} for more information about $\alpha_i$. In Eq.~\ref{eq:aerobat-full-dynamics}, $D_u$, $D_{ua}$, and $H_u$ are the block matrices from partitioning Aerobat's full-dynamics. For partitioning, the full-dynamics inertia matrix $D(q)$, Coriolis matrix $C(q,\dot q)$ and conservative potential forces (gravity and rubber bands) $G(q)$ corresponding to the underactuated (position and orientation) $q_u=[p_A^\top,q_A^\top]^\top$ and actuated (the wings joints) $a(t)$ are separated. 

The force and moments $f_e$ and $m_e$ in Eq.~\ref{eq:body-force-moment} are obtained as elastic conservative forces. To do this, the total potential energy $V(q)$ is used to obtain $H=C(q,\dot q)\dot q + G(q)$ in Aerobat's model.  $V(q)$ is given given by
\begin{equation}
    V(q)=\frac{1}{2}K(p_G-p_A)^\top(p_G-p_A) \\+ m_Ag\Big(p_{G,z}+R_A^0p_{A,z}\Big)
    \label{eq:total-potential-energy}
\end{equation}
\noindent where $K$ denotes the rubber bands' elastic coefficient and $m_A$ is Aerobat's total mass. Note that in Eq.~\ref{eq:aerobat-full-dynamics}, We model aerobat-fluidic environment interactions using our experimentally tested unsteady-quasi-static models that are evaluated at discrete strips along the wing span. The output from this aerodynamic model $y=[y1,\dots,y_m]^\top$ where $y_i$ is the external aerodynamic force at i-th strip is governed by the state-space model $\dot \xi$. I briefly cover this model; however, the reader is referred to \cite{sihite2022unsteady} for more details on the model derivations. 

\subsection{Brief Overview of Unsteady Aerodynamic State-Space Model $\dot \xi=A(\xi)\xi+B(\xi)a(t)$}

We superimpose horseshoe vortices on and behind the wing blade elements to calculate lift and drag forces. Consider the time-varying circulation value at $s_i$, the location of the $i$-th element on the wing, denoted by $\Gamma_i(t)$. The circulation can be parameterized by truncated Fourier series of $n$ coefficients. $\Gamma_i$ is given by 
\begin{equation}
\begin{aligned}
    \Gamma_i(t) = a^\top(t)
    \begin{bmatrix}
    \sin(\theta_i)\\
    \vdots\\
    \sin(n\theta_i)\\
    \end{bmatrix},
\end{aligned}
    \label{eq:circulation}
\end{equation}
\noindent where $a=[a_1,\dots,a_n]^\top$ are the Fourier coefficients and $\theta_i=\arccos(\frac{s_i}{l})$ ($l$ is the wingspan size). From Prandtl's lifting line theory, additional circulation-induced kinematics denoted by $y_{\Gamma}$ are considered on all of the points $p_i$. These circulation-induced kinematics are given by
\begin{equation}
\begin{aligned}
    y_{\Gamma} = 
    \begin{bmatrix}
    1&\frac{\sin{2\theta_1}}{\sin{\theta_1}}&\dots&\frac{\sin{n\theta_1}}{\sin{\theta_1}}\\
    1&\frac{\sin{2\theta_2}}{\sin{\theta_2}}&\dots&\frac{\sin{n\theta_2}}{\sin{\theta_2}}\\
    \vdots\\
    1&\frac{\sin{2\theta_n}}{\sin{\theta_n}}&\dots&\frac{\sin{n\theta_n}}{\sin{\theta_n}}\\
    \end{bmatrix}a(t)
    \label{eq:induced-kin}
\end{aligned}
\end{equation}
Now, we utilize a Wagner function $\Phi(\tau) = \Sigma_{k=1}^2\psi_k \exp{(-\frac{\epsilon_k}{c_i} \tau)}$, where $\psi_k$, $\epsilon_k$ are some scalar coefficients and $\tau$ is a scaled time to compute the aerodynamic force coefficient response $\beta_{i}$ associated with the i-th blade element. The kinematics of i-th element using Eqs.~\ref{eq:aerobat-full-dynamics} and \ref{eq:induced-kin} are given by 
\begin{equation}
y'_{1,i} = y_{1,i} + y_{\Gamma,i}.    
\end{equation}
\noindent Following Duhamel's integral rule, the response is obtained by the convolution integral given by
\begin{equation}
    \beta_i = y'_{1,i}\Phi_0 + \int_0^t\frac{\partial \Phi(t-\tau)}{\partial \tau}y'_{1,i}d\tau \\
\label{eq:lift_coeff_wagner}
\end{equation}
\noindent where $\Phi_0=\Phi(0)$. We perform integration by part to eliminate $\frac{\partial \Phi}{\partial \tau}$, substitute the Wagner function given above in \eqref{eq:lift_coeff_wagner}, and employ the change of variable given by $z_{k,i} (t) = \int_{0}^{t} \exp{(-\frac{\epsilon_k}{c_i}(t-\tau))} y'_{1,i} d\tau$, where $k\in\{1,2\}$, to obtain a new expression for $\beta_i$ based on $z_{k,i}$  
\begin{equation}
\beta_i =
y'_{1,i}\Phi_0+
\begin{bmatrix}
\psi_1\frac{\epsilon_1}{c_i}&\psi_2\frac{\epsilon_2}{c_i}
\end{bmatrix}
\begin{bmatrix}
z_{1,i}\\
z_{2,i}
\end{bmatrix}
\label{eq:?1}
\end{equation}
The variables $a$, $z_{1,i}$, and $z_{2,i}$ are used towards obtaining a state-space realization that can be marched forward in time. Using the Leibniz integral rule for differentiation under the integral sign, unsteady Kutta–Joukowski results $\beta_i=\frac{\Gamma_i}{c_i}+\frac{d\Gamma_i}{dt}$, and \eqref{eq:circulation}, the model that describes the time evolution of the aerodynamic states is obtained
\begin{equation}
\Sigma_{Aero,i}:\left\{
\begin{aligned}
    A_i\dot{a} &= -B_ia + C_iZ_i  + \Phi_0y'_{1,i}\\
    \dot{Z}_i &= D_iZ_i + E_iy'_{1,i}\\
\end{aligned}
\right.
    \label{eq:aerodynamic-ss-model}
\end{equation}
\noindent where $Z_i$, $A_i$, $B_i$, $C_i$, $D_i$, and $E_i$ are the state variable and matrices corresponding to the i-th blade element. They are given by
\begin{equation}
\begin{aligned}
    Z_i &= 
    \begin{bmatrix}
    z_{1,i} & z_{2,i}
    \end{bmatrix}^\top,\\
    A_i &= 
    \begin{bmatrix}
    \sin{\theta_i} & \sin{2\theta_i} & \dots & \sin{n\theta_i} 
    \end{bmatrix},\\
    B_i &= A_i/c_i,\\
    C_i &= 
    \begin{bmatrix}
    \frac{\psi_1\epsilon_1}{c_i} & \frac{\psi_2\epsilon_2}{c_i}
    \end{bmatrix},\\
    D_i &= 
    \begin{bmatrix}
    \frac{-2\epsilon_1}{c_i} & 0\\
    0 & \frac{-2\epsilon_2}{c_i}
    \end{bmatrix},\\
    E_i &=
    \begin{bmatrix}
    2-\exp{\frac{\epsilon_1 t}{c_i}} & 2-\exp{\frac{\epsilon_2 t}{c_i}}
    \end{bmatrix}^\top.
\end{aligned}
    \label{eq:?2}
\end{equation}
\noindent We define the unified aerodynamic state vector, used to describe the state space of \eqref{eq:aerobat-full-dynamics}, as following $\xi = [a^\top, Z^\top]^\top$, where $Z=[Z^\top_1,\dots,Z^\top_n]^\top$.

\section{Control}

The full dynamics of the guard with Aerobat sitting inside are given by
\begin{equation}
\Sigma_{FullDyn}\left\{\begin{array}{l}
\dot{x}_1=x_2 \\
\dot{x}_2=g_1+g_2u+g_3x_3 \\
\dot{x}_3=G(t) \\
z=x_1
\end{array}\right.
    \label{eq:extended-state-model}
\end{equation}
\noindent where $x_1=[p_G^\top,q_G^\top]^\top$, the nonlinear terms $g_i$ are given by Eqs.~\ref{eq:guard-modell} and \ref{eq:aerobat-full-dynamics}, $u=[\dots f_i \dots]$ from Eq.~\ref{eq:body-force-moment}, and $x_3=y$ from Eq.~\ref{eq:aerobat-full-dynamics}. As it can be seen, the model is extended with another state $x_3$ because $y$ (aerodynamic force) can be written in the state-space form given by Eq.~\ref{eq:aerobat-full-dynamics}. 

Now, the control of Eq.~\ref{eq:extended-state-model} is considered here, assuming that $u$ is calculated only based on $z=x_1$ observations. The time-varying term $G(t)$ (the dynamics of $y$) is highly nonlinear; however, we have an efficient model of $G(t)$ \cite{sihite2022unsteady}. We obtained an aerodynamic model in \cite{sihite2022unsteady} that closely predicts the external aerodynamic forces impinged on Aerobat. 

Using this model, we establish a state observer for $x_3$ to augment the feedback $u=Kx_2$ (where $K$ is the control gain) so that $\dot x_2$ remains bounded and stable. Consider the following definition of estimated states $\hat x_i$ from \cite{sihite2022unsteady}:
\begin{equation}
\begin{aligned}
& \hat{x}_1=\hat{x}_2-\beta_1\left(\hat{x}_1-x_1\right) \\
& \hat{x}_2=g_1+g_2 u+g_3\hat{x}_3-\beta_2\left(\hat{x}_1-x_1\right) \\
& \hat{x}_3=-\beta_3\left(\hat{x}_1-x_1\right)
\end{aligned}
    \label{eq:estimated-states}
\end{equation}
\noindent where $\beta_i$ is the observer gains. Now, we define the error $e_i=\hat x_i- x_1$ for $i=1,2,3$. The following observer model is found
\begin{equation}
\left[\begin{array}{c}
\dot e_1 \\
\dot e_2\\
\dot e_3
\end{array}\right]=\left[\begin{array}{ccc}
-\beta_1 & I & 0 \\
-\beta_2 & 0 & g_3 \\
-\beta_3 & 0 & 0
\end{array}\right]\left[\begin{array}{c}
e_1 \\
e_2 \\
e_3
\end{array}\right]+\left[\begin{array}{c}
0 \\
0 \\
-I
\end{array}\right] G(t)
    \label{eq:observer}
\end{equation}
\noindent The gains $\beta_i$ for the observer given by Eq.~\ref{eq:observer} can be obtained if upper bounds for $G$ and $g_2$ can be assumed. We have extensively studied $g_1$, $g_2$ and $g_3$ terms in Aerobat's model in past and ongoing efforts. Based on the bounds for $\|g_1\|$, $\|g_2\|$, and $\|g_3\|$, I tuned the observer. The controller used for the bounding flight is given by
\begin{equation}
u = g_2^{-1}\Big(u_0 -g_1 - g_3\hat x_3\Big) 
    \label{eq:controller}
\end{equation}
\noindent where $u_0=Kx_2$.

\section{Perception}

I tested perception and feature tracking on Aerobat using ORBSLAM3, and data collection using Aerobat's onboard hardware. Due to hardware restrictions only data collection was run on onboard hardware and ORBSLAM3 was post-processed on a linux machine. ORBSLAM3 represents a state-of-the-art Simultaneous Localization and Mapping (SLAM) system that supports multiple sensor configurations including monocular, stereo, and RGB-D cameras with visual-inertial capabilities \cite{noauthor_orb-slam3_nodate, campos_orb-slam3_2021}. The system employs Oriented FAST and Rotated BRIEF (ORB) features for real-time feature detection and matching, enabling robust performance across various environments.

\begin{figure*}[h!]
\vspace{0.08in}
    \centering
    \includegraphics[width=\linewidth]{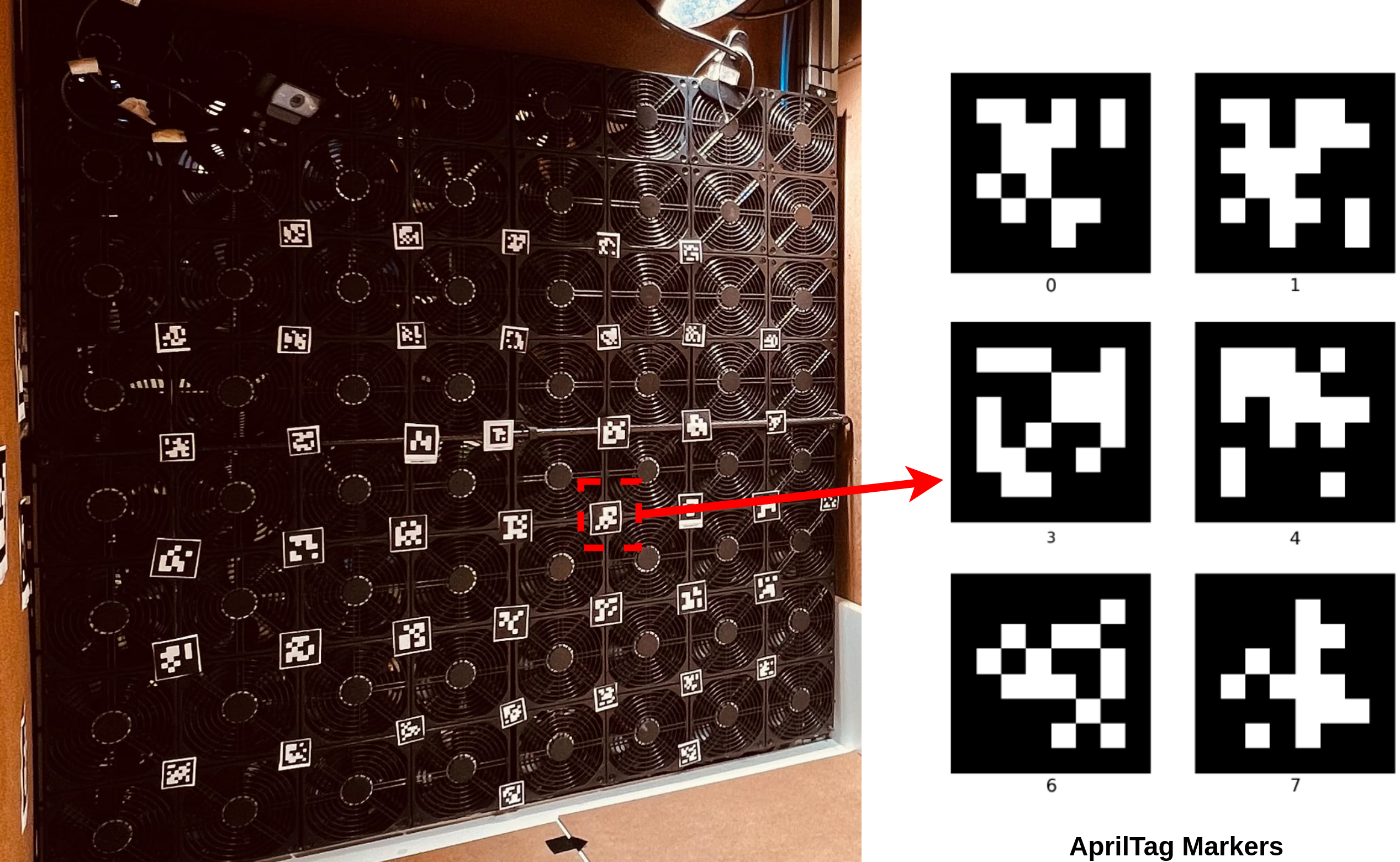}
    \caption{AprilTag Features for ORBSLAM detection in RISE Arena}
    \label{fig:orbslam_rise}
\vspace{-0.08in}
\end{figure*}

\begin{figure*}[h!]
\vspace{0.08in}
    \centering
    \includegraphics[width=0.8\linewidth]{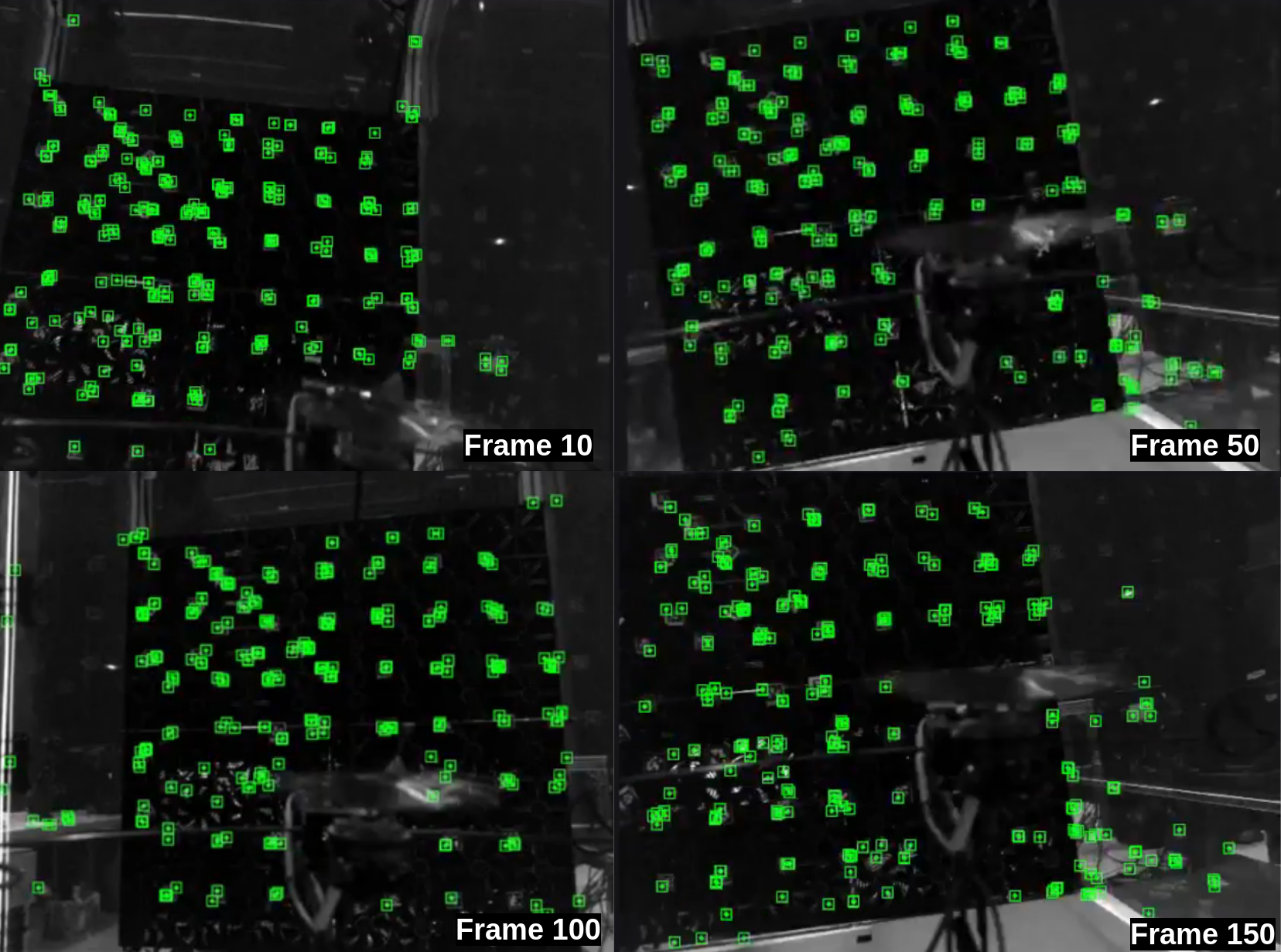}
    \caption{ORBSLAM Feature Detection on multiple frames while Aerobat is flapping}
    \label{fig:orbslam_feature}
\vspace{-0.08in}
\end{figure*}

The implemented setup utilized a Raspberry Pi Zero coupled with a Raspberry Pi Camera and an Adafruit ICM20948 IMU. This configuration presented specific challenges due to hardware limitations and synchronization issues. The system operated with the IMU sampling at 200 Hz and the camera capturing frames at 20 Hz, creating a fundamental timing disparity. For better feature detection we installed Apriltag markers on the Fan Array of the Robotics-Inspired Study and Experimentation(RISE) Arena Fig. ~\ref{fig:orbslam_rise}, which resulted in more accurate feature detection of all the features as shown in Fig. ~\ref{fig:orbslam_feature}
\\
\\
I tested two different modes for ORBSLAM: 
\begin{enumerate}
\item In monocular mode, ORB-SLAM3 operates using only camera input, performing feature extraction and matching across consecutive frames to estimate camera motion and build a sparse 3D map. This mode demonstrates robust performance in well-lit environments with sufficient texture, achieving centimeter-level precision in ideal conditions.

\item The mono-inertial configuration combines visual features with IMU data, requiring non-constant acceleration and angular velocity for proper initialization \cite{noauthor_mono-inertial_nodate}. This mode typically offers enhanced robustness compared to pure monocular operation, particularly during rapid movements or challenging lighting conditions \cite{noauthor_robustness_nodate}.
\end{enumerate}

A critical limitation in the current implementation was the lack of hardware synchronization between the IMU and camera sensors. The 200 ms timing offset between sensors significantly impacted system performance, leading to frequent map resets. This timing discrepancy particularly affected the system's ability to maintain consistent tracking during extended operation periods.

\begin{figure*}[h!]
\vspace{0.08in}
    \centering
    \includegraphics[width=0.8\linewidth]{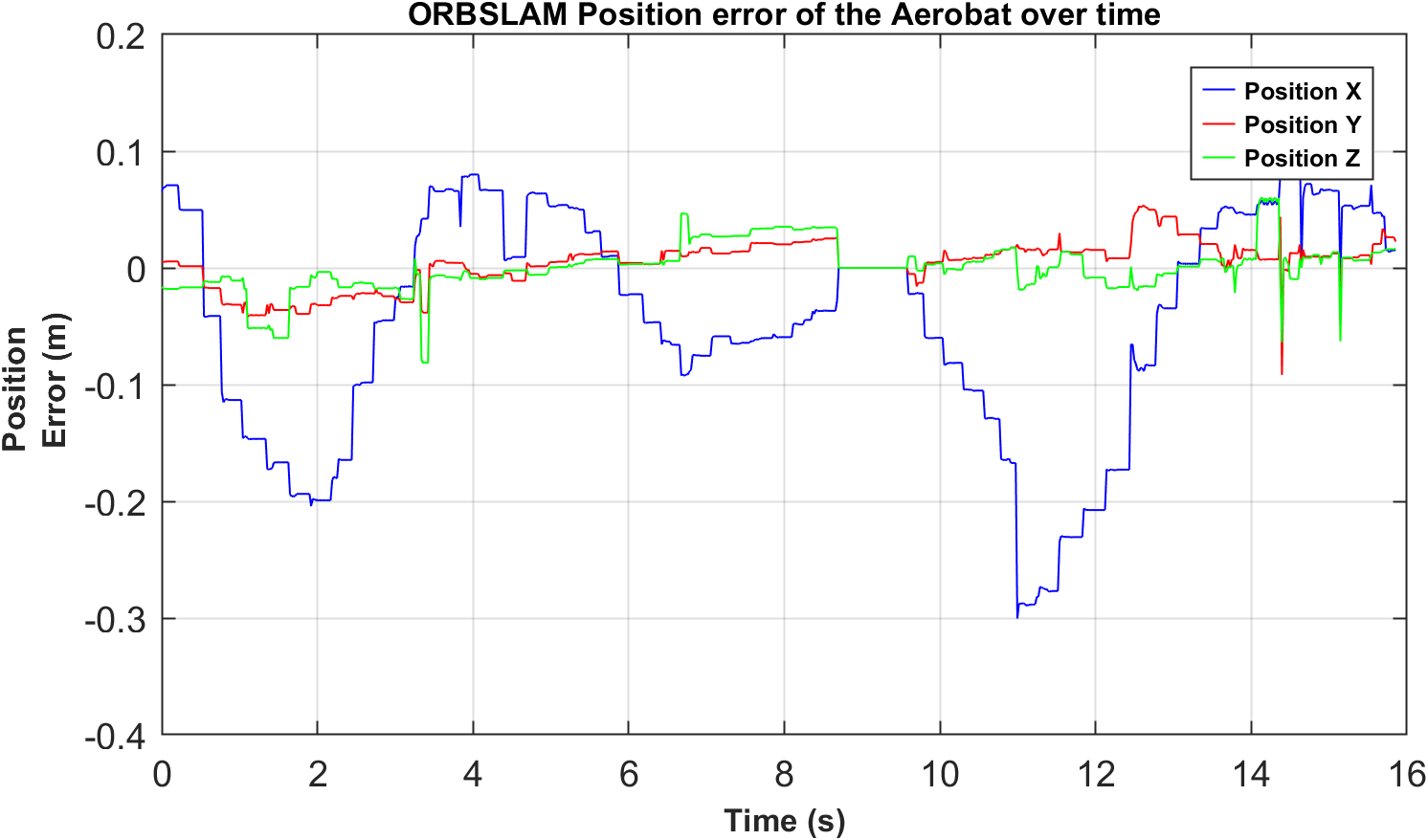}
    \caption{Position Estimation of Aerobat while flapping using ORBSLAM3}
    \label{fig:orbslam_pos}
\vspace{-0.08in}
\end{figure*}

\begin{figure*}[h!]
\vspace{0.08in}
    \centering
    \includegraphics[width=0.8\linewidth]{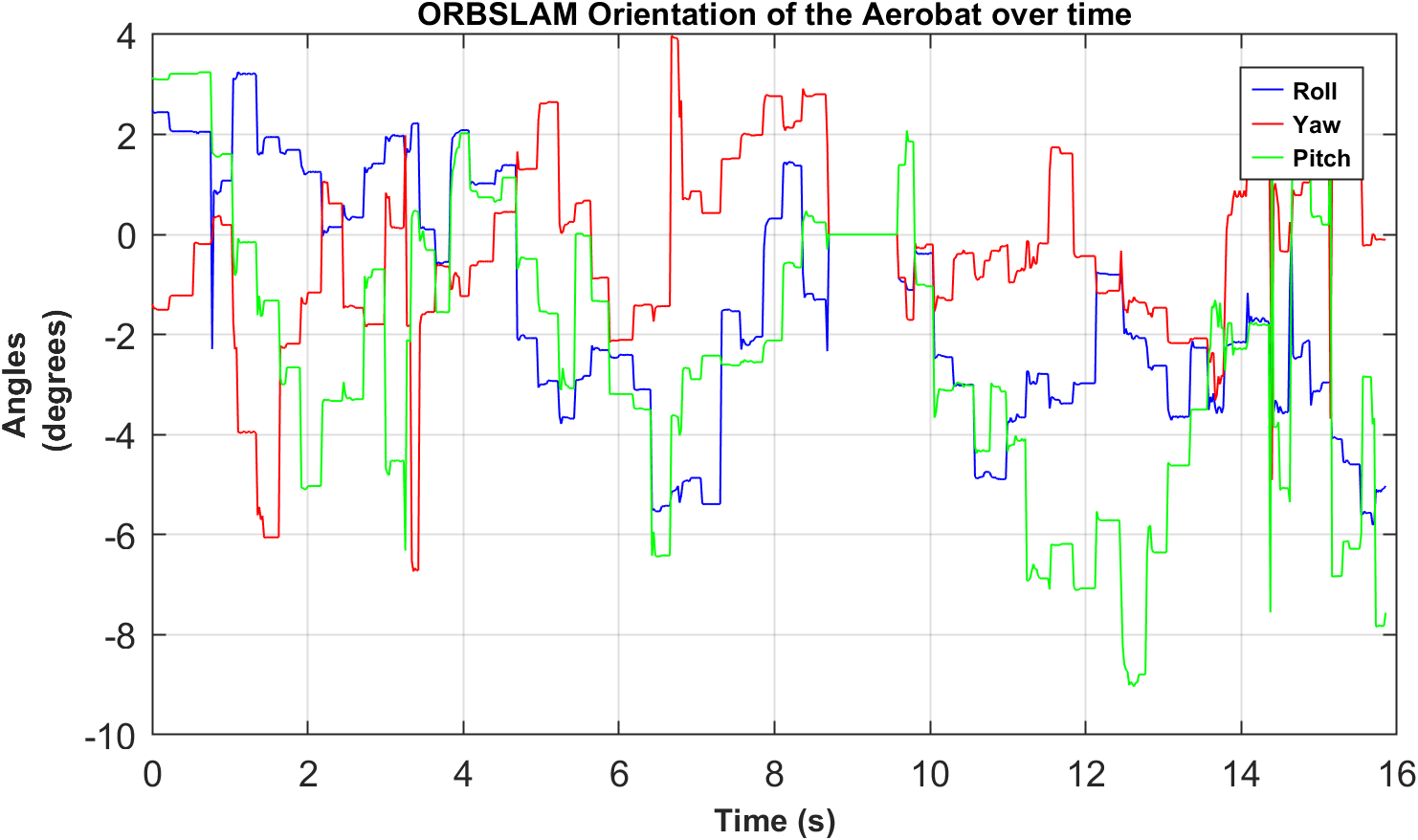}
    \caption{Orientation Estimation of Aerobat while flapping using ORBSLAM3}
    \label{fig:orbslam_ori}
\vspace{-0.08in}
\end{figure*}

Despite synchronization challenges, the system demonstrated capability in short-duration tests (approximately 10 seconds) Fig.~\ref{fig:orbslam_pos} ~\ref{fig:orbslam_ori} shows the estimated position and orientation using ORBSLAM, we can see that it oscillates in X axis, its mostly due to the flapping while down and upstroke causes extreme noise in that axis. The results showed meaningful mapping and tracking, though with notable noise introduction due to the temporal misalignment of sensor data. The system achieved centimeter-precision tracking during stable periods before map resets occurred \cite{noauthor_robustness_nodate}.

The implementation highlights the necessity for hardware-level synchronization, specifically recommending the use of a time-synchronized shutter camera and IMU. This upgrade would eliminate the current 200 ms delay between sensors and significantly improve system stability and mapping consistency. Additionally, the integration of global optimization techniques could further enhance the system's robustness against temporal misalignments \cite{wang_robust_2023, xiao_sl-slam_2024, noauthor_240503413v3_nodate}.




\section{Experimentation}

\subsection{Test Setup}

The Robotics-Inspired Study and Experimentation (RISE) arena provides a controlled testing environment for the Aerobat Guard system, featuring precise motion tracking and artificial wind generation capabilities. The facility is enclosed by transparent acrylic panels for easier expirementations. The dimesions of this arena are 63 x 40 x 60 inches. 

The arena's motion capture system consists of six OptiTrack cameras mounted strategically around the upper perimeter. A key feature of the RISE arena is its wind generation system, comprising a 10×10 array of sectionally controlled fans which enables the simulation of various aerodynamic conditions with precise control over wind speed. The experimental setup includes an 8V regulated power supply system for the Aerobat Guard, delivering stable power through a tethered connection. This configuration allows for extended flight testing without battery constraints while maintaining the freedom of movement necessary for aerobatic maneuvers. The entire system is monitored and controlled through a dedicated workstation running the motion capture software and control algorithms.

\begin{figure}[h!]
\vspace{0.08in}
    \centering
    \includegraphics[width=1\linewidth]{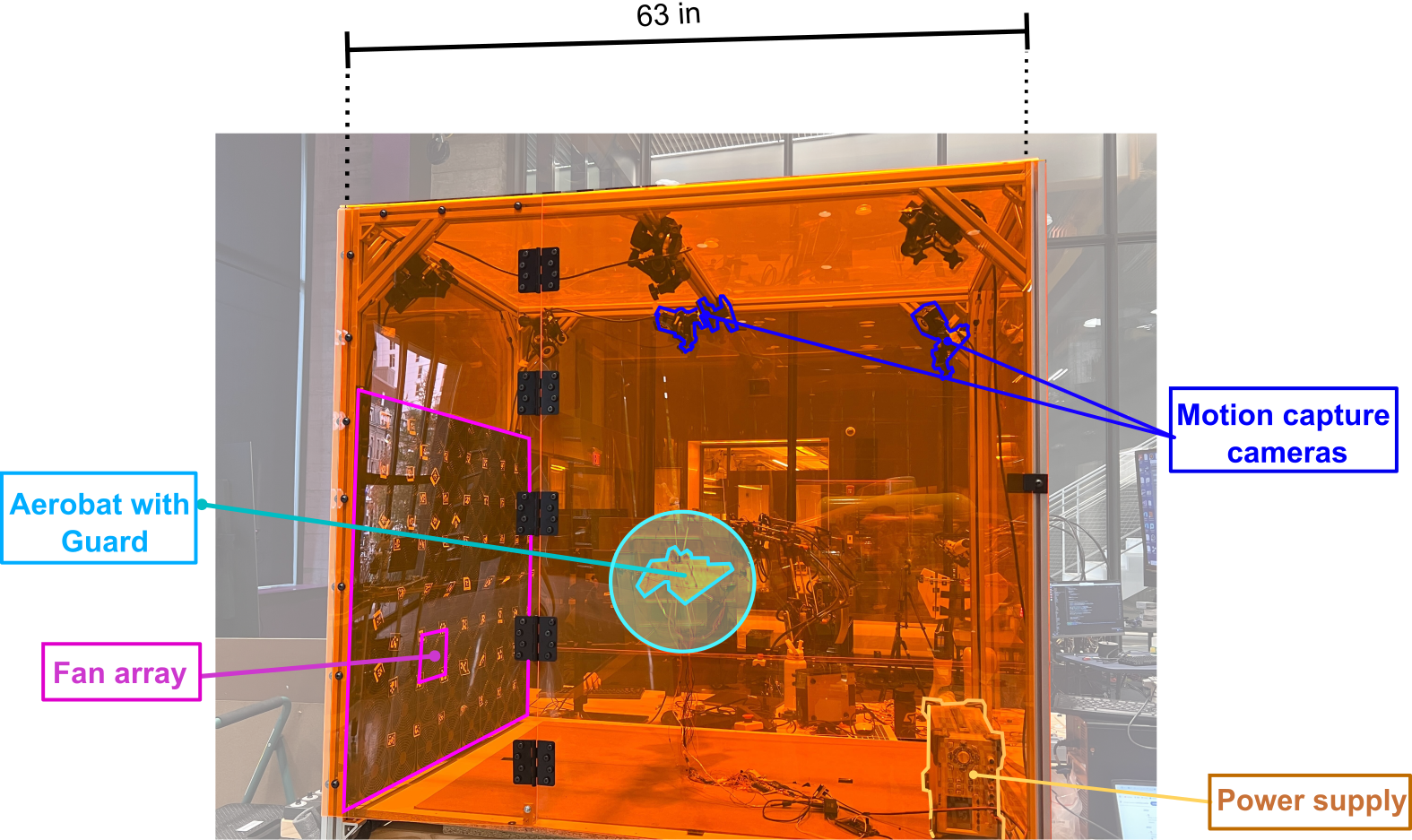}
    \caption{Robotics-Inspired Study and Experimentation (RISE) Arena at Northeastern University}
    \label{fig:rise_arena}
\vspace{-0.08in}
\end{figure}

\subsection{Tuning of Controllers}

The Aerobat Guard system employs a cascaded control architecture with two distinct PID controllers operating at 200 Hz. The Position PID controller, running on the Linux computer, manages the global position and orientation tracking based on OptiTrack motion capture feedback. The Attitude PID controller, implemented on the ESP32 microcontroller, handles local stability using IMU measurements for roll and pitch corrections.

The tuning process begins with the inner loop Attitude PID controller, where gains are adjusted with the Aerobat Guard in a constrained setup to prevent unstable oscillations. Initial tuning focuses on achieving stable hover with minimal drift, starting with conservative proportional gains and gradually introducing derivative terms to dampen oscillations. The integral terms are then carefully adjusted to eliminate steady-state errors while avoiding integral windup.

\begin{figure*}
\vspace{0.08in}
    \centering
    \includegraphics[width=0.8\linewidth]{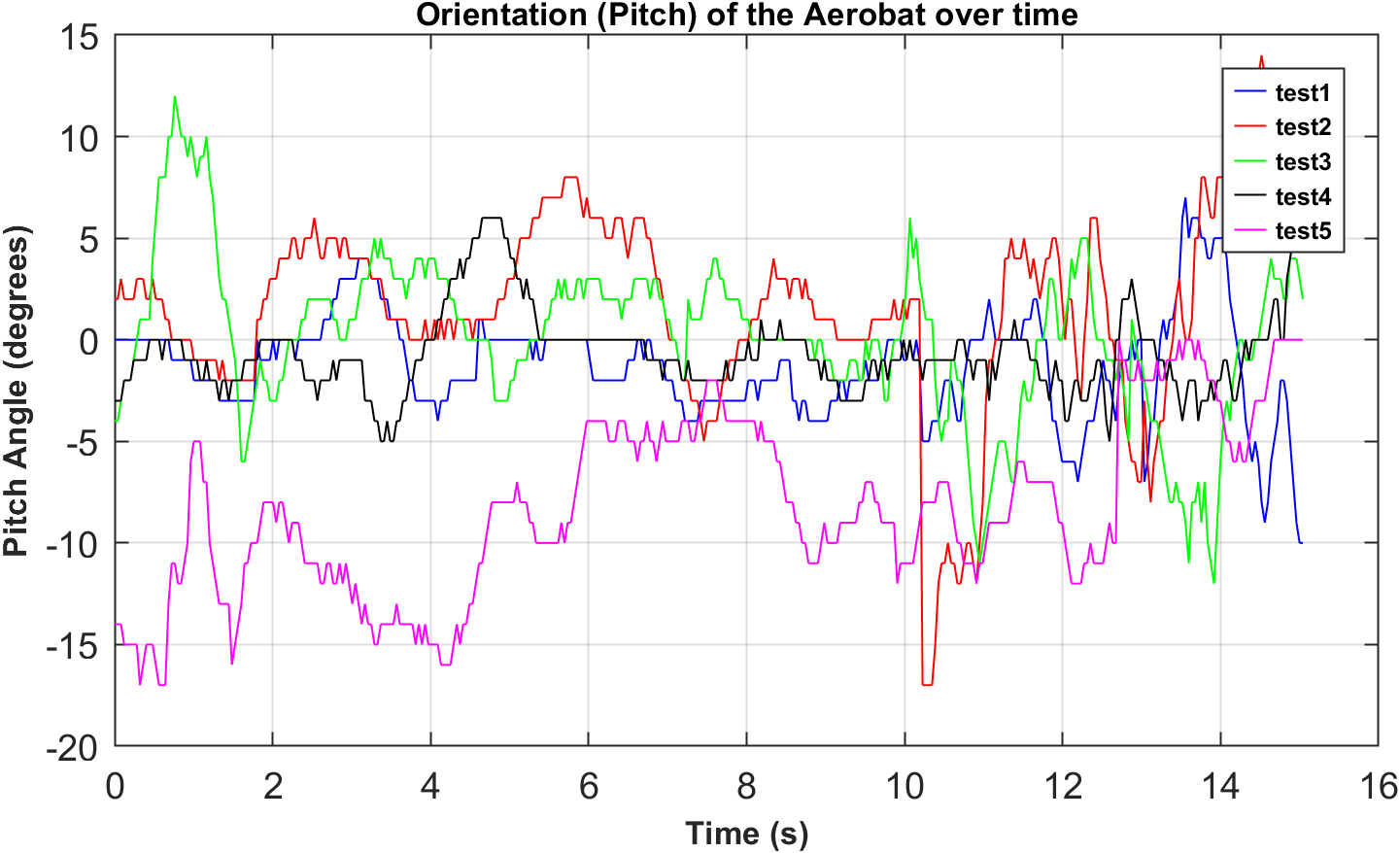}
    \hfill
    
    \caption{Shows the pitch over different tests}
    \label{fig:pitch_tests}
\vspace{-0.08in}
\end{figure*}

\begin{figure*}
\vspace{0.08in}
    \centering
    \includegraphics[width=0.8\linewidth]{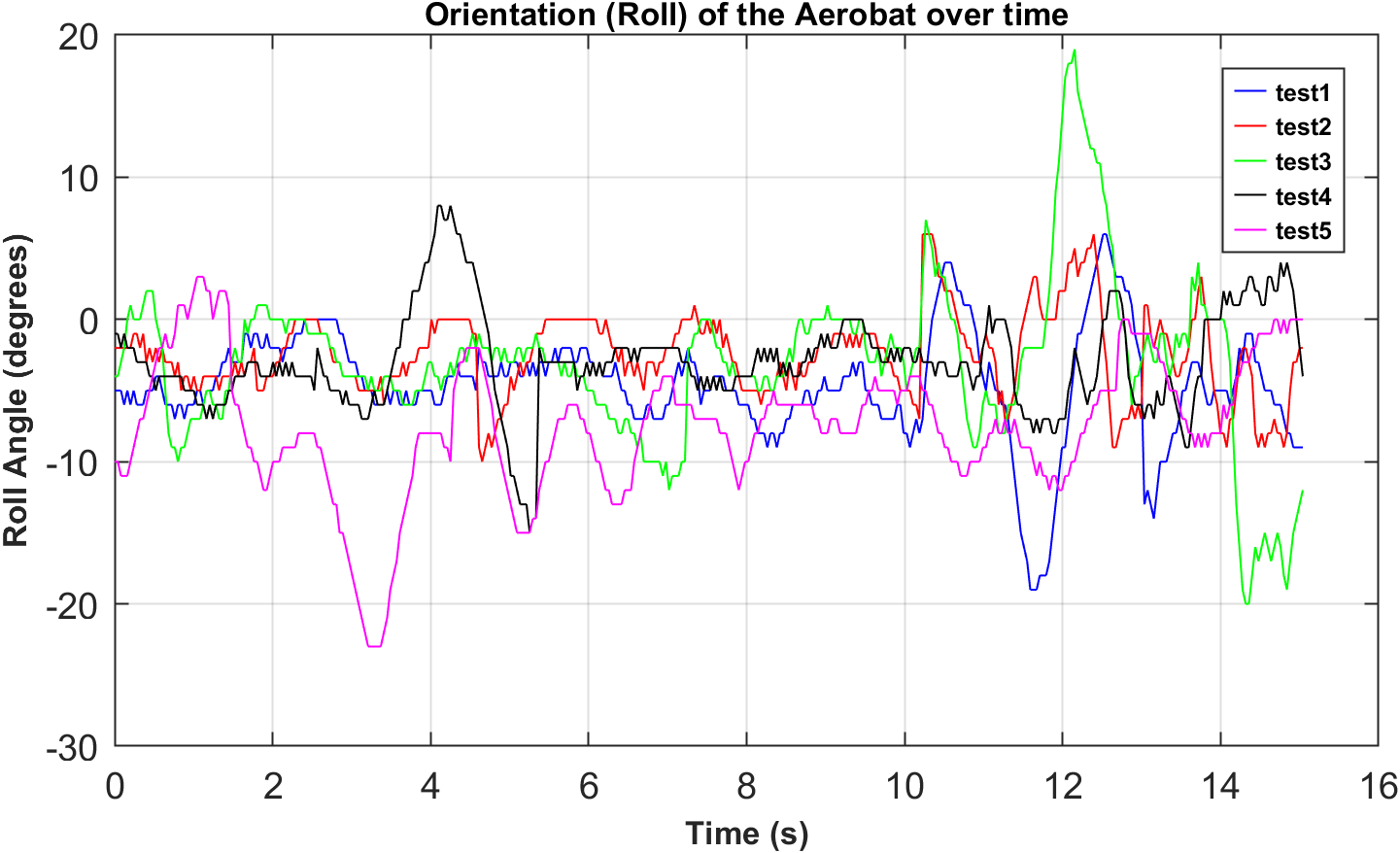}
    \hfill
    
    \caption{Shows the roll over different tests}
    \label{fig:roll_tests}
\vspace{-0.08in}
\end{figure*}

\begin{figure*}
\vspace{0.08in}
    \centering
    \includegraphics[width=0.8\linewidth]{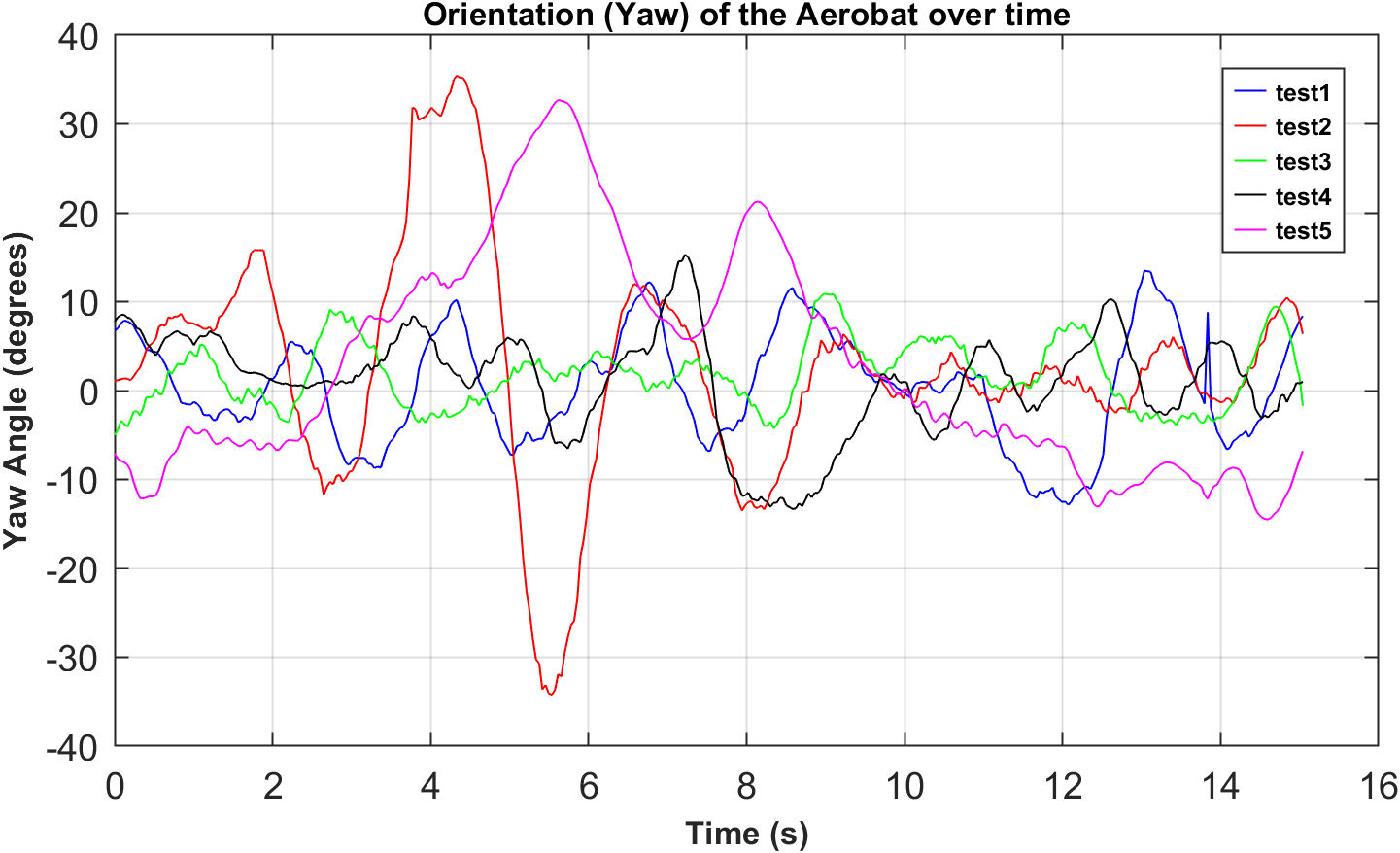}
    \hfill
    
    \caption{Shows the yaw over different tests}
    \label{fig:yaw_tests}
\vspace{-0.08in}
\end{figure*}

After stabilizing the attitude control, the outer loop Position PID controller is tuned. This process starts with the Aerobat Guard in a hover state near ground level, typically 1 meter height. The position controller gains are incrementally increased while monitoring the system's response to small position setpoint changes. The tuning prioritizes smooth transitions and minimal overshoot, particularly critical for the safety-oriented guard structure.

The initial flight state requires careful consideration of the guard's unique geometry and mass distribution. The system is typically initialized in a stable hover configuration, with all six motors running at balanced thrust levels to maintain level orientation. The guard's protective structure, while providing safety benefits, introduces additional aerodynamic considerations that must be accounted for in the control gain selection, particularly during aggressive maneuvers or in the presence of external disturbances.

\begin{figure*}
\vspace{0.08in}
    \centering
    \includegraphics[width=0.48\linewidth]{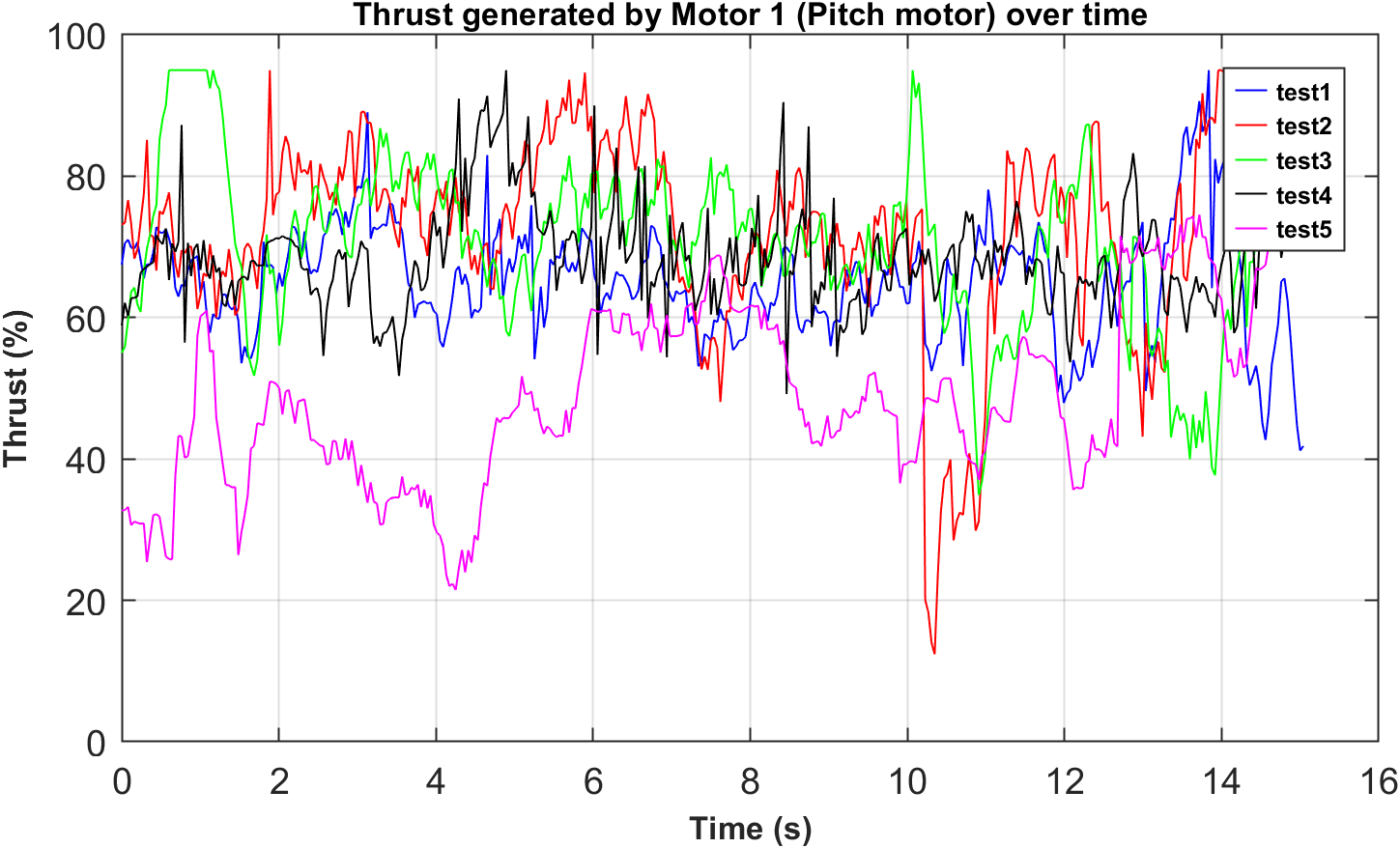}
    \hfill
    \includegraphics[width=0.48\linewidth]{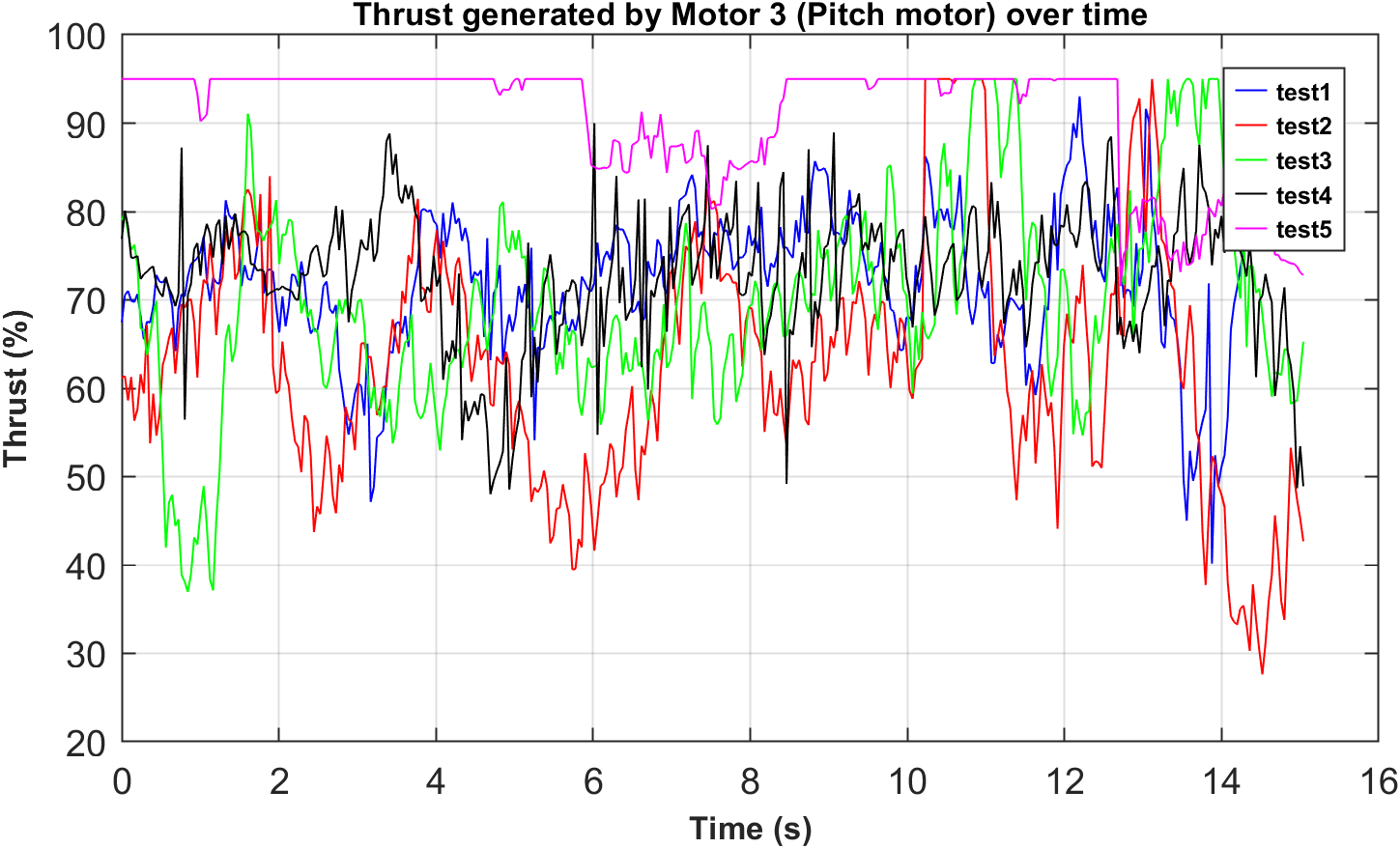}
    \caption{Shows the thrust generated by pitch motors $f_1$ and $f_3$ over different tests}
    \label{fig:pitch_thrust}
\vspace{-0.08in}
\end{figure*}

\begin{figure*}
\vspace{0.08in}
    \centering
    \includegraphics[width=0.48\linewidth]{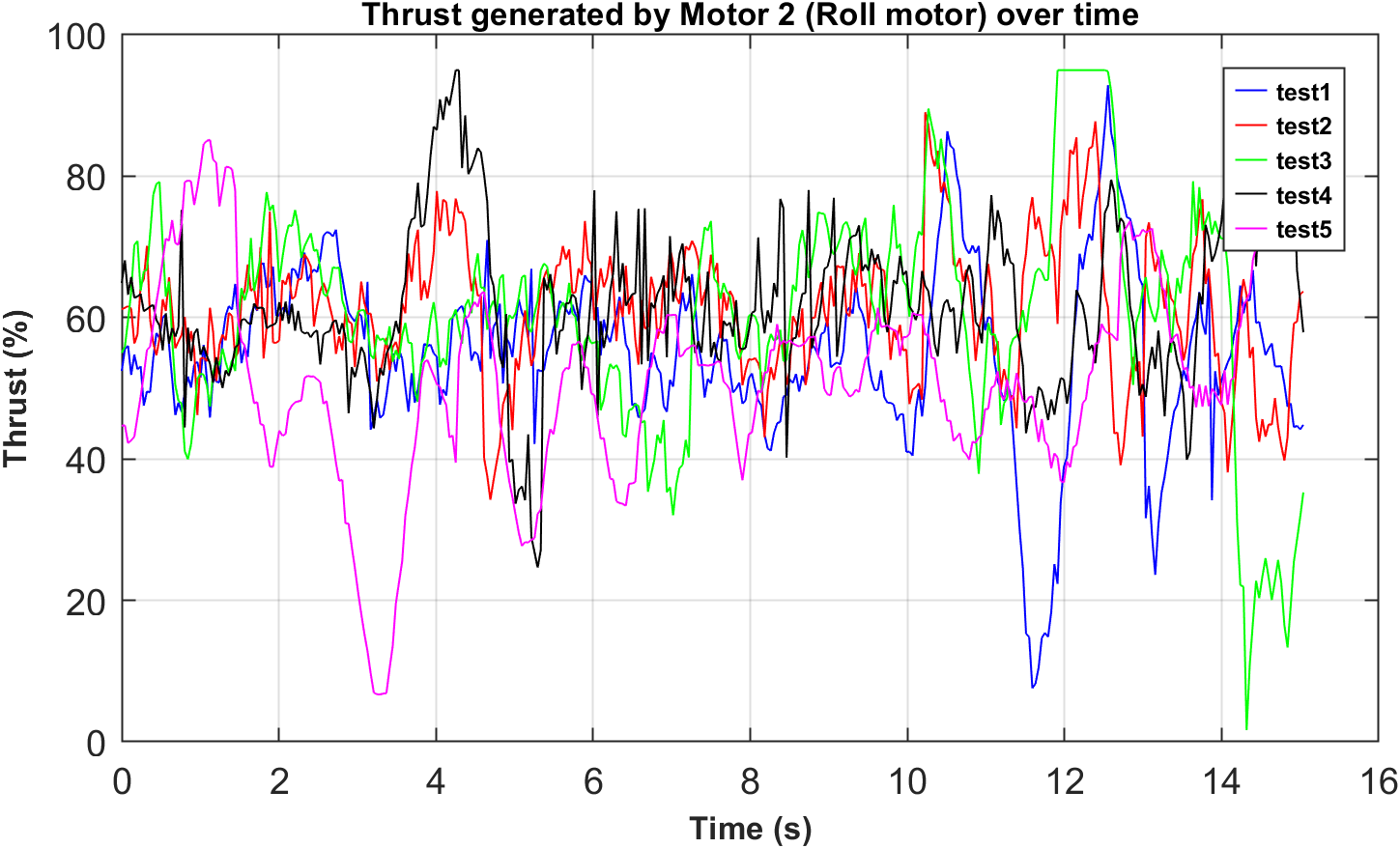}
    \hfill
    \includegraphics[width=0.48\linewidth]{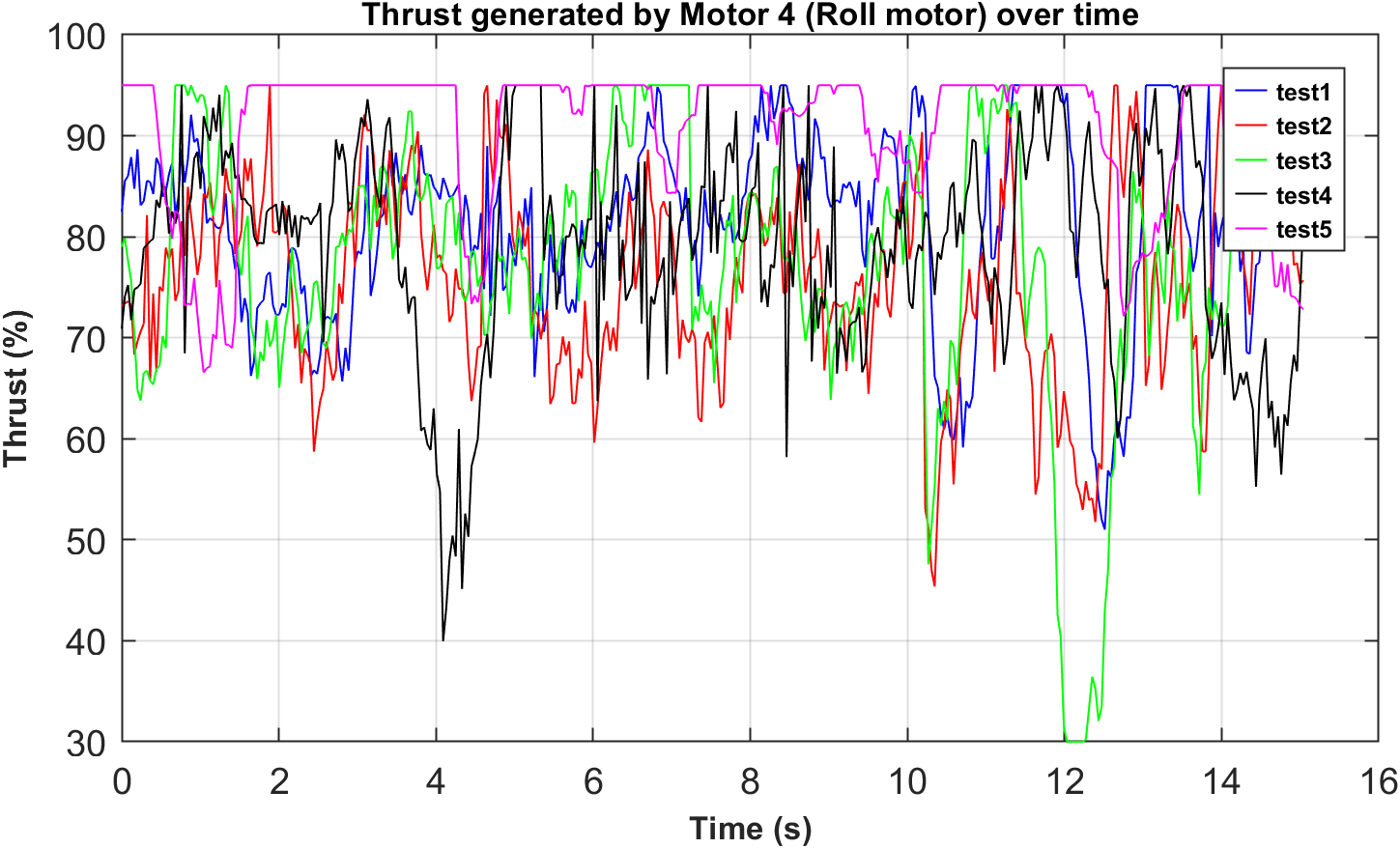}
    \caption{Shows the thrust generated by roll motors $f_2$ and $f_4$ over different tests}
    \label{fig:roll_thrust}
\vspace{-0.08in}
\end{figure*}

\begin{figure*}
\vspace{0.08in}
    \centering
    \includegraphics[width=0.48\linewidth]{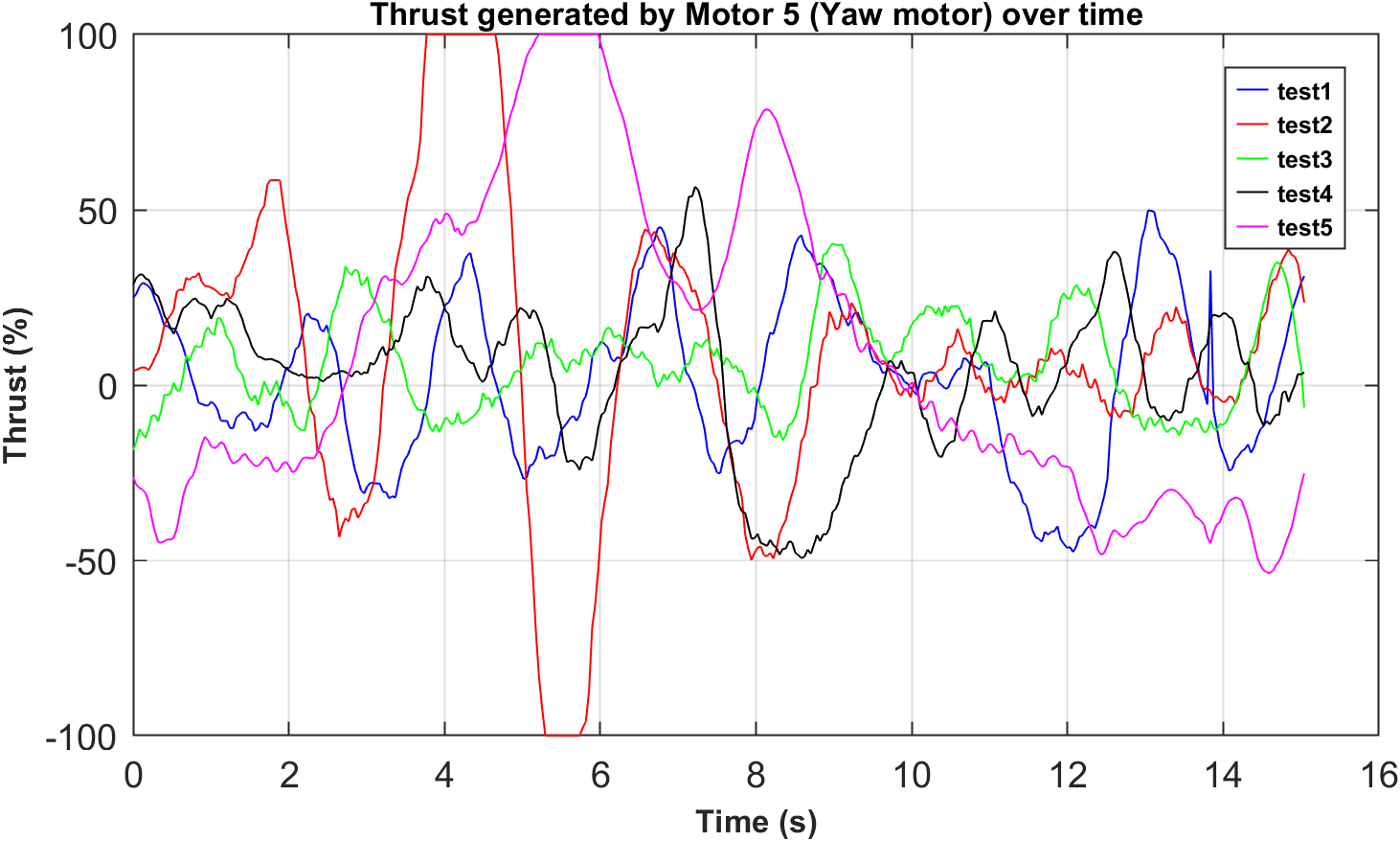}
    \hfill
    \includegraphics[width=0.48\linewidth]{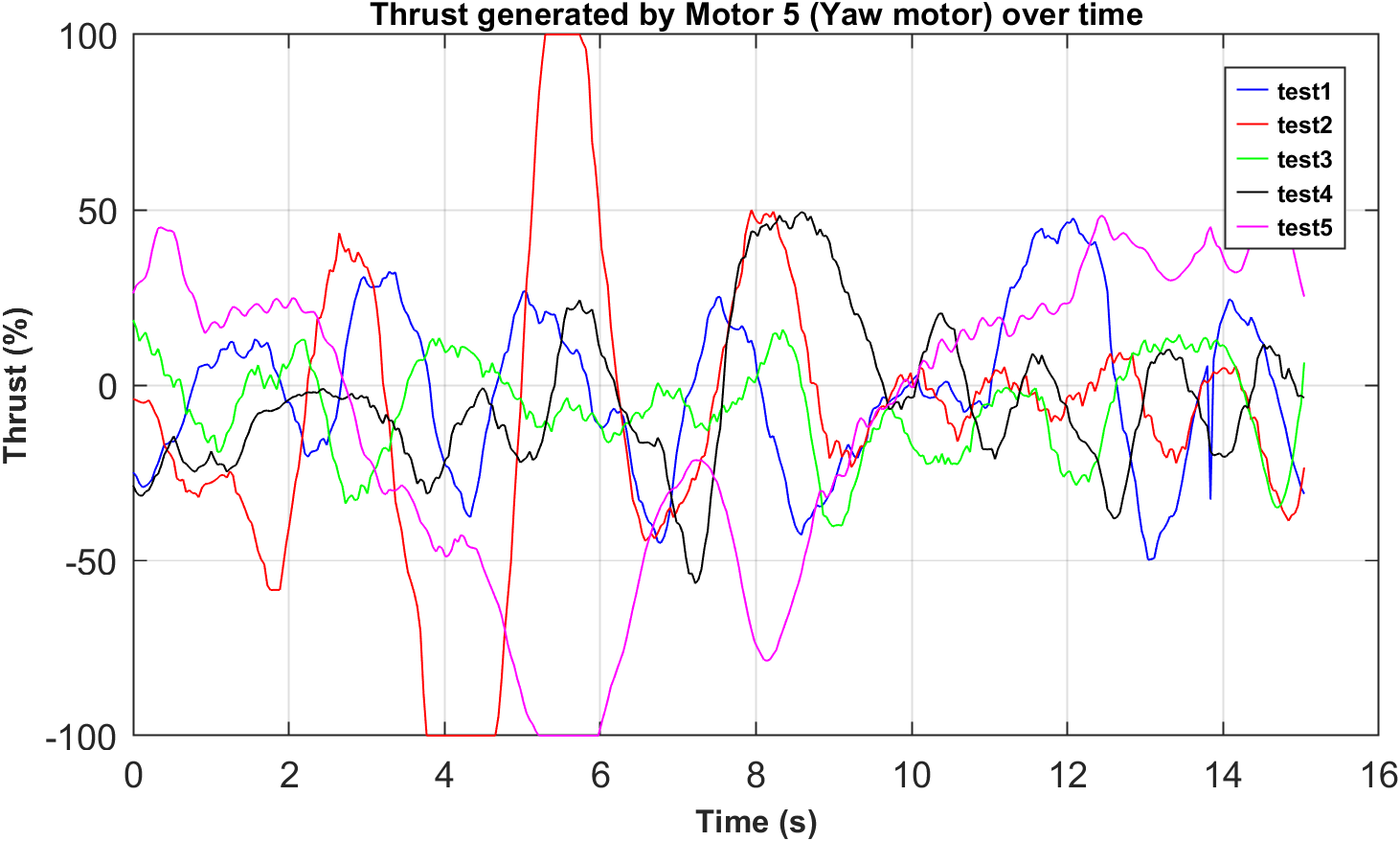}
    \caption{Shows the thrust generated by yaw motors $f_5$ and $f_6$ over different tests}
    \label{fig:yaw_thrust}
\vspace{-0.08in}
\end{figure*}

This section presents a comparative analysis of five distinct Position PID controller configurations for a three-axis position stabilization for Aerobat guard. The experiments were conducted with varying gain parameters while maintaining a constant Z-axis setpoint of 0.2 meters.

\begin{table}[h]
\caption{X-Axis Position PID Gains Comparison}
\begin{tabular}{|c|c|c|c|c|c|}
\hline
Test & P-Gain & I-Gain & D-Gain & P/I Ratio & P/D Ratio \\
\hline
1 & 15.900 & 0.300 & 31.000 & 53.000 & 0.513 \\
2 & 15.900 & 0.300 & 31.000 & 53.000 & 0.513 \\
3 & 18.264 & 0.648 & 40.691 & 28.205 & 0.449 \\
4 & 18.727 & 0.535 & 39.305 & 34.987 & 0.476 \\
5 & 17.900 & 0.450 & 36.000 & 39.778 & 0.497 \\
\hline
\end{tabular}
\label{Tab:x_gains}
\end{table}

\begin{table}[h]
\caption{Y-Axis Position PID Gains Comparison}
\begin{tabular}{|c|c|c|c|c|c|}
\hline
Test & P-Gain & I-Gain & D-Gain & P/I Ratio & P/D Ratio \\
\hline
1 & 15.900 & 0.300 & 35.000 & 53.000 & 0.454 \\
2 & 15.900 & 0.300 & 35.000 & 53.000 & 0.454 \\
3 & 17.617 & 0.631 & 36.304 & 27.919 & 0.485 \\
4 & 18.028 & 0.497 & 35.000 & 36.261 & 0.515 \\
5 & 15.900 & 1.282 & 35.000 & 12.398 & 0.454 \\
\hline
\end{tabular}
\label{Tab:y_gains}
\end{table}

\begin{table}[h]
\caption{Z-Axis Position PID Gains Comparison}
\begin{tabular}{|c|c|c|c|c|c|}
\hline
Test & P-Gain & I-Gain & D-Gain & P/I Ratio & P/D Ratio \\
\hline
1 & 36.000 & 3.500 & 30.000 & 10.286 & 1.200 \\
2 & 36.000 & 3.500 & 30.000 & 10.286 & 1.200 \\
3 & 36.000 & 3.500 & 30.000 & 10.286 & 1.200 \\
4 & 36.000 & 3.500 & 30.000 & 10.286 & 1.200 \\
5 & 36.000 & 3.500 & 30.000 & 10.286 & 1.200 \\
\hline
\end{tabular}
\label{Tab:z_gains}
\end{table}

\begin{table}[h]
\caption{Error Analysis Summary}
\begin{tabular}{|c|c|c|c|c|}
\hline
Test & RMS Error X & RMS Error Y & RMS Error Z & Total RMS \\
\hline
1 & 3.670 & 2.460 & 0.085 & 4.419 \\
2 & 4.509 & 3.157 & 0.093 & 5.505 \\
3 & 4.374 & 3.085 & 0.075 & 5.354 \\
4 & 4.406 & 2.835 & 0.090 & 5.240 \\
5 & 3.783 & 7.334 & 0.076 & 8.252 \\
\hline
\end{tabular}
\label{Tab:total_rms}
\end{table}

\subsubsection{X-Axis Position Control Performance}
The X-axis controller demonstrated significant variations in gain configurations across tests. Notable observations include:

\begin{itemize}
    \item Tests 1 and 2 employed identical conservative gains ($K_p=15.900$, $K_i=0.300$, $K_d=31.000$)
    \item A systematic increase in proportional gain was observed in Tests 3-5, reaching a maximum of $K_p=18.727$ in Test 4
    \item The P/I ratio exhibited a decreasing trend from 53.000 to approximately 30-40 in later tests, indicating enhanced integral action
    \item The derivative gain showed substantial variation, ranging from 31.000 to 40.691
\end{itemize}

\subsubsection{Y-Axis Position Control Performance}
The Y-axis controller exhibited unique characteristics:

\begin{itemize}
    \item Base configuration remained consistent with X-axis for Tests 1-2
    \item Test 5 implemented a significantly higher integral gain ($K_i=1.282$), resulting in a markedly lower P/I ratio of 12.398
    \item Derivative gain maintained relative stability at 35.000, with minor variations
    \item The varying P/I ratios suggest different approaches to steady-state error minimization
\end{itemize}

\subsubsection{Z-Axis Position Control Performance}
Z-axis control parameters demonstrated remarkable consistency:

\begin{itemize}
    \item Uniform gain values maintained across all tests ($K_p=36.000$, $K_i=3.500$, $K_d=30.000$)
    \item Higher proportional gain compared to X and Y axes
    \item Elevated integral gain indicating enhanced steady-state error correction
    \item Balanced P/D ratio of 1.200 suggesting optimal damping characteristics
\end{itemize}

\subsection{Error Analysis}
Table \ref{Tab:total_rms} presents the RMS error metrics for each axis. Key findings include:

\begin{itemize}
    \item Test 1 achieved optimal performance with the lowest total RMS error of 4.419
    \item Z-axis control demonstrated consistent precision with RMS errors ranging from 0.075 to 0.093
    \item Y-axis control in Test 5 exhibited significant degradation (RMS error = 7.334)
    \item X-axis performance remained relatively consistent across all tests (RMS error range: 3.670-4.509)
\end{itemize}

\subsection{Performance Metrics}
The comprehensive performance analysis revealed:

\subsubsection{RMS Error}
The Root Mean Square error for each axis is calculated as:

\begin{equation}
\text{RMS}_{\text{axis}} = \sqrt{\frac{1}{n}\sum_{i=1}^{n}(x_i - x_{\text{target}})^2}
\end{equation}

where $x_{\text{target}}$ is 0.2m for Z-axis and 0m for X and Y axes. The total RMS error combines all axes:

\begin{equation}
\text{RMS}_{\text{total}} = \sqrt{\text{RMS}_x^2 + \text{RMS}_y^2 + \text{RMS}_z^2}
\end{equation}

\subsubsection{Stability Metric}
Position consistency is quantified using:

\begin{equation}
\text{Stability Metric} = -\frac{1}{3}(\sigma_x + \sigma_y + \sigma_z)
\end{equation}

where $\sigma_x$, $\sigma_y$, $\sigma_z$ are position standard deviations. The final performance score combines both metrics:

\begin{equation}
\text{Performance Score} = -\text{RMS}_{\text{total}} + \text{Stability Metric}
\end{equation}

Results indicate:
\begin{itemize}
    \item Test 1: Superior performance (Score: -4.509)
    \item Tests 2-4: Moderate performance (Scores: -5.336 to -5.620)
    \item Test 5: Suboptimal performance (Score: -8.342)
\end{itemize}

This experimental analysis yields several significant findings:

\begin{enumerate}
    \item Conservative gain configurations (Test 1) demonstrated superior overall performance
    \item Increased integral action did not necessarily correlate with improved performance metrics
    \item Z-axis control exhibited robust stability across all configurations
    \item Higher P/I ratios generally corresponded to enhanced system performance
    \item Y-axis control showed particular sensitivity to integral gain adjustments
\end{enumerate}

\begin{figure*}
\vspace{0.08in}
    \centering
    \includegraphics[width=0.8\linewidth]{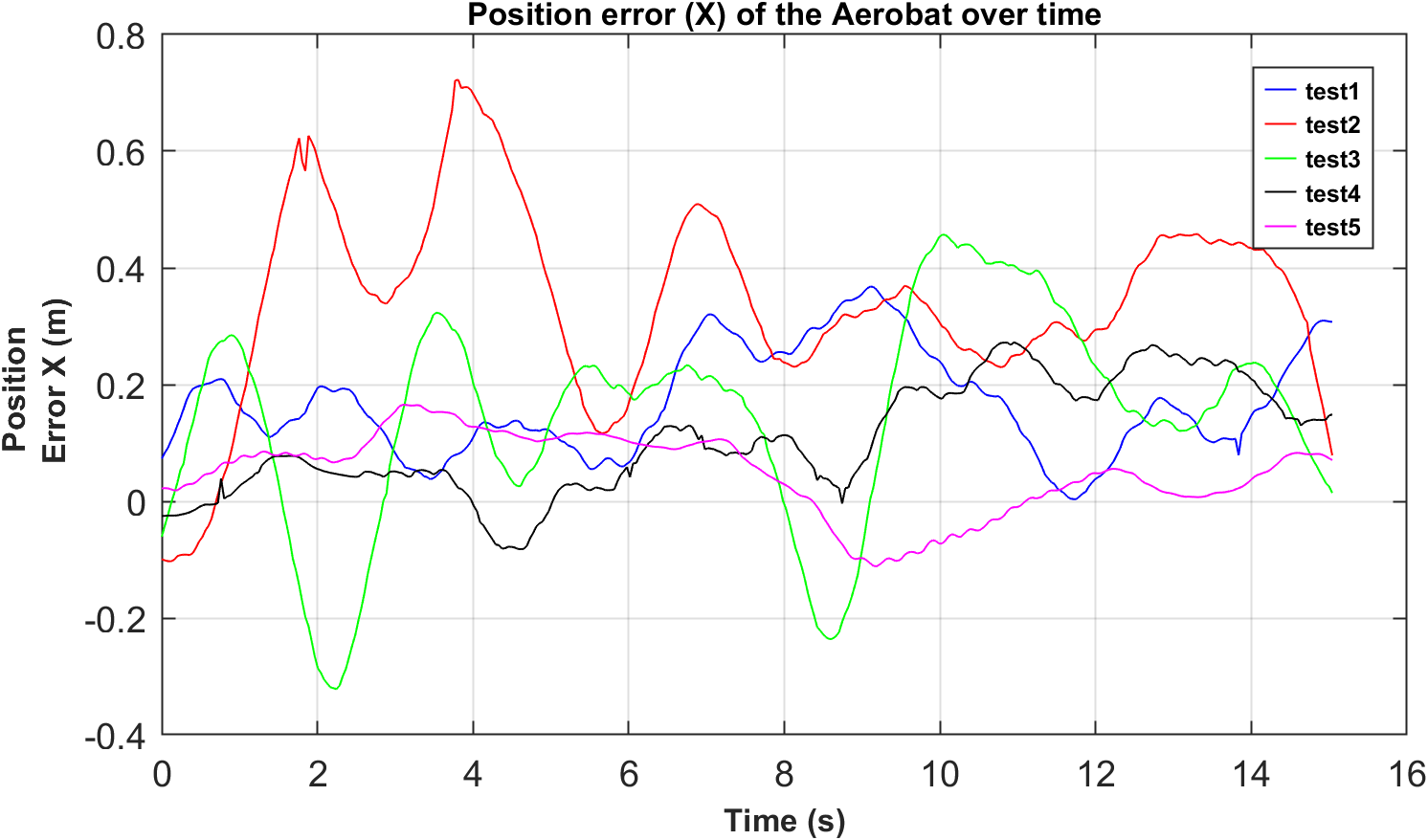}
    \caption{Position Error in X Axis}
    \label{fig:pos_err_x}
\vspace{-0.08in}
\end{figure*}

\begin{figure*}
\vspace{0.08in}
    \centering
    \includegraphics[width=0.8\linewidth]{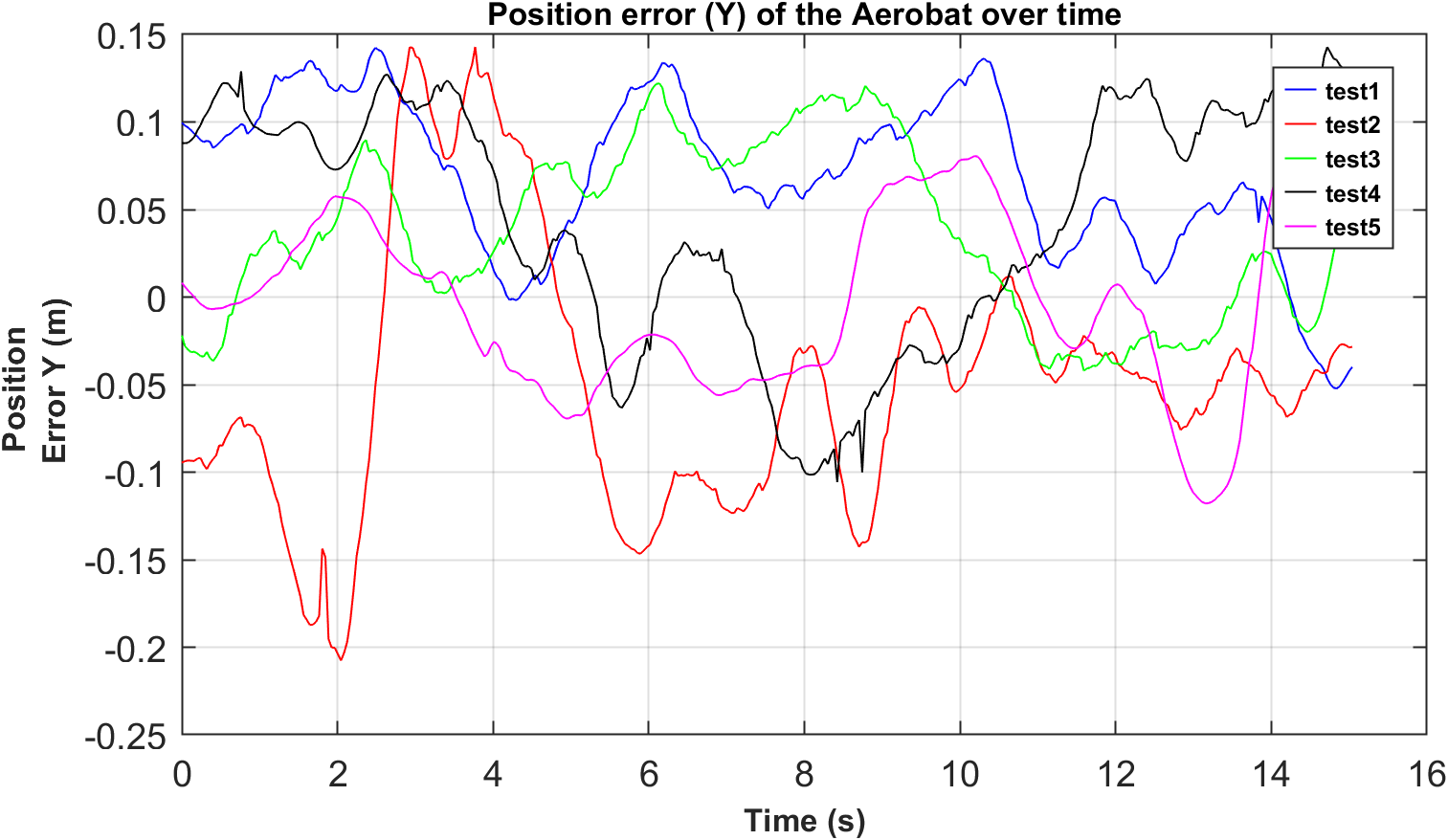}
    \caption{Position Error in Y Axis}
    \label{fig:pos_err_y}
\vspace{-0.08in}
\end{figure*}

\begin{figure*}
\vspace{0.08in}
    \centering
    \includegraphics[width=0.8\linewidth]{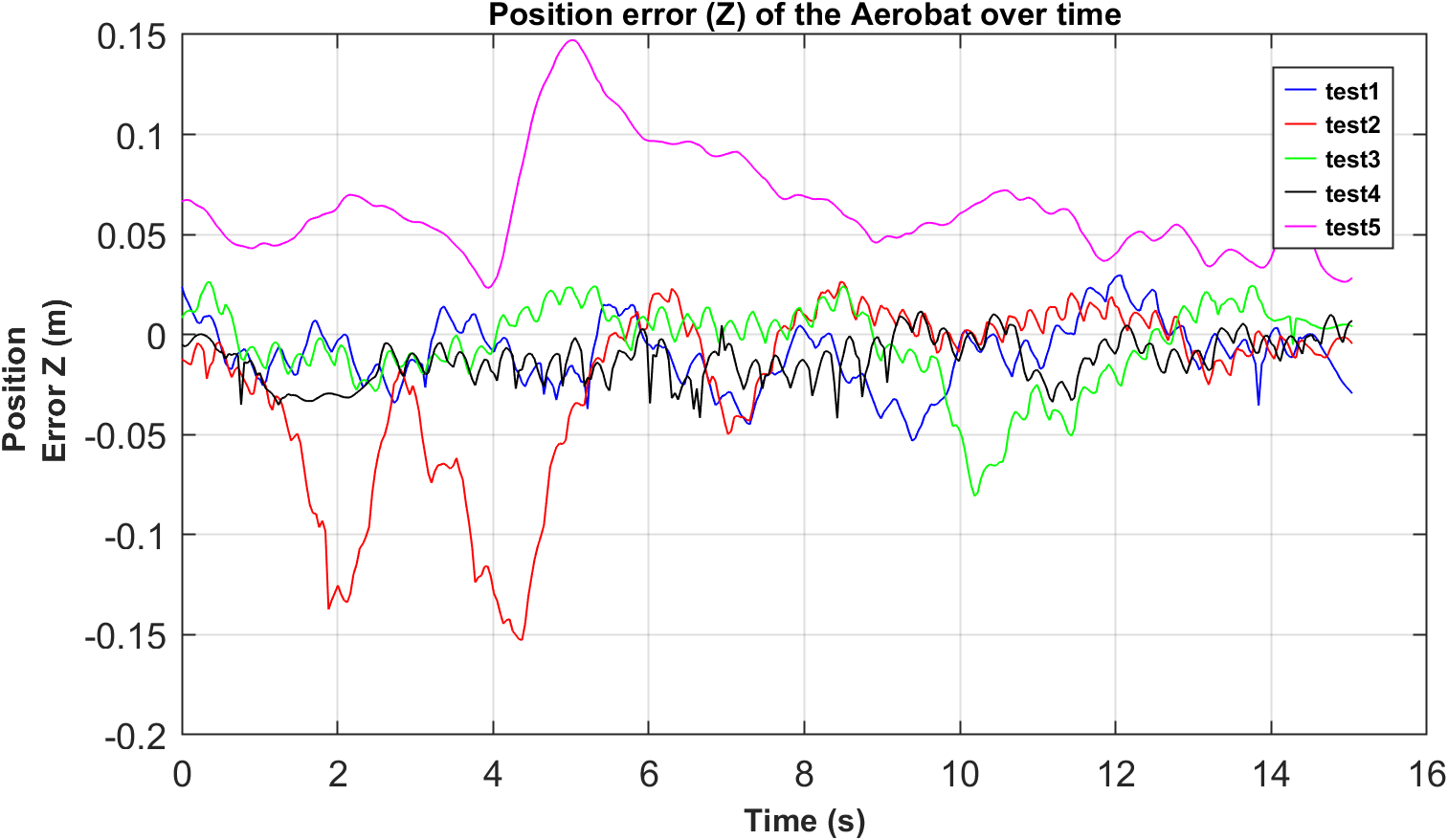}
    \caption{Position Error in Z Axis}
    \label{fig:pos_err_z}
\vspace{-0.08in}
\end{figure*}

These results suggest that optimal performance in this three-axis system is achieved through conservative gain selection with appropriate balance between proportional and integral action. The findings provide valuable insights for future tuning strategies in similar control applications.

\section{Results}

\begin{figure}
\vspace{0.08in}
    \centering
    \includegraphics[width=1\linewidth]{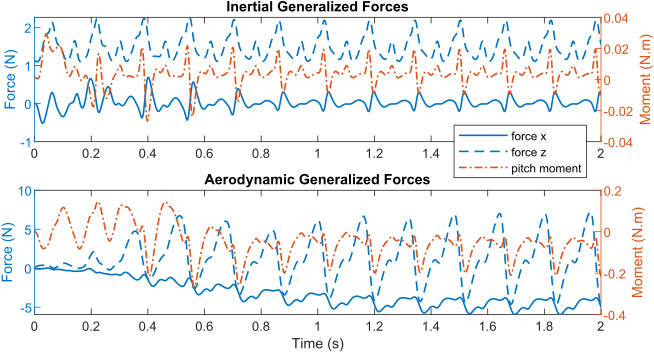}
    \caption{Shows estimated generalized inertial $g_2^{-1}g_1$ and aerodynamics $g_2^{-1}g_3\hat x_3$ contributions based on knowledge on the boundedness of \textbf{$\|g_2\|$} and \textbf{$\|G(t)\|$}.}
    \label{fig:gen_force}
\vspace{-0.08in}
\end{figure}

The calculation of the control command  $u=[f_1,\dots,f_6]^\top$ was achieved based on estimated values of the extended state $x_3$ in hovering inside our lab. The individual thrust values are given in Figs. ~\ref{fig:pitch_thrust}~\ref{fig:roll_thrust} ~\ref{fig:yaw_thrust}, $x_2$, which embodies the guard's world position and orientations and velocities, were measured using my OptiTrack system and the onboard inertial measurement unit. Our unsteady aerodynamic model reported in \cite{sihite2022unsteady} was used to identify the bounds on $G(t)$ during hovering. Based on unsteady aerodynamic model results, for hovering, $\|G(t)\|$ was selected. The estimated values $g_1^{-1}g_2\hat x_3$ for hovering denoted as generalized aerodynamic force contributions are shown in Fig.~\ref{fig:gen_force}. The estimated values for $g_1^{-1}f$ denoted as generalized inertial dynamics contributions and shown in Fig.~\ref{fig:gen_force} were used to complete the computation of $u$. The performance of the controller in stabilizing the roll, pitch, yaw, x-y-z positions are shown in Figs.~\ref{fig:pitch_tests} ~\ref{fig:roll_tests} ~\ref{fig:yaw_tests} ~\ref{fig:pos_err_x}~\ref{fig:pos_err_y} ~\ref{fig:pos_err_z} .


\section{Concluding Remarks}

Aerobat, in addition to being tailless, possesses morphing wings that add to the inherent complexity of flight control. In this work, I covered my efforts to stabilize the flight dynamics of Aerobat. We employed a guard design with manifold small thrusters to stabilize Aerobat's position and orientation in hovering. We developed a dynamic model of Aerobat and guard interacting with each other. For flight control purposes, we assumed the guard cannot observe Aeroat's states. Then, we proposed an observer design to estimate the unknown states of the Aerobat-guard dynamics, which were used for closed-loop hovering control. I reported culminating experimental and tuning results that showcase the effectiveness of my approach. 

\nocite{sihite_unsteady_2022,sihite_unilateral_2021-1, ramezani_generative_2021-1, lessieur_mechanical_2021, de_oliveira_thruster-assisted_2020, grizzle_progress_nodate, sihite_multi-modal_2023, sihite_orientation_2021, ramezani_towards_2020}

\chapter{Harpy: Thruster-Assisted Bipedal Locomotion}
\label{chap:harpy}

\section{Introduction}

Raibert's robots \cite{murphy_littledog_2011} and those from Boston Dynamics \cite{noauthor_robots_nodate} stand out as some of the most successful legged robots, capable of robust hopping or trotting even amid substantial unplanned disturbances. In recent years, there have also been numerous successful bipedal dynamic walker designs from various research groups and institutions.

We have seen bipedal robots push beyond human athletic capabilities, as exemplified by Atlas, which performs acrobatics such as jumping, flipping, parkour, running, bounding, and gymnastic routines. Despite these remarkable abilities, all such systems are fundamentally limited by their operation principles. These robots regulate contact forces through posture manipulation, constrained by underactuation and the unilaterality of contact forces, which imposes significant limitations.

\begin{figure}[t]
    \vspace{0.05in}
    \centering
    \includegraphics[width=0.5\textwidth]{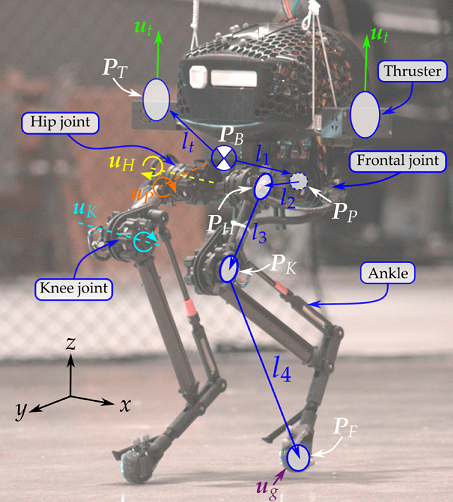}
    \caption{Illustrates the Harpy platform, which motivates our thruster-assisted dynamic legged locomotion.}
    \label{fig:cover-image}
    \vspace{-0.05in}
\end{figure}

A key objective of the Harpy project (shown in Fig.~\ref{fig:cover-image}), a collaboration with Northeastern University and Caltech, is to surpass the capabilities of standard bipedal robots by devising means to manipulate contact forces through both posture manipulation and thrust vectoring. This goal is deeply rooted in our understanding of locomotion principles in resilient biological systems like birds, which are known for their cursorial advancements \cite{abourachid_bird_2011}. For example, Chukar birds can use their flapping wings to generate external forces that help them manipulate contact forces when traversing steep slopes \cite{dial_wing-assisted_2003-1}.

Birds' ability to achieve agile legged locomotion, beyond what terrestrial systems can typically achieve, is due to their synchronous use of legs and wings. Robotic biomimicry of this behavior at the hardware and control levels presents significant challenges. Integrating dynamic walking and aerial mobility is a particularly difficult hardware problem, as it requires reconciling conflicting design requirements for each mode of operation.

\begin{figure}[h]
  \centering
    \includegraphics[width=0.7\textwidth]{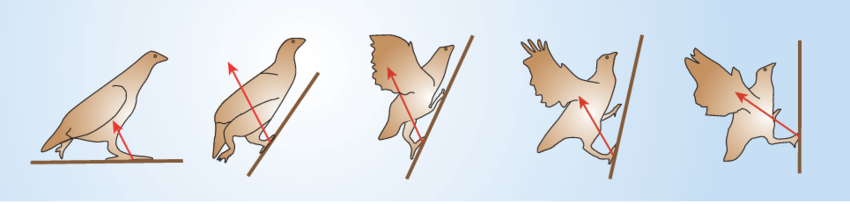}
  \caption[Chukar Bird Inclined Walking]{Wing assisted walking of Chukar Bird \cite{john_r_hutchinson_pdf_nodate}}
  \label{fig:chukar_inclined}
\end{figure}

Recent designs have explored the application of thrusters (i.e., thrust vectoring) and posture manipulation in notable robots such as the Multi-modal Mobility Morphobot (M4) \cite{sihite_multi-modal_2023,sihite_efficient_2022,mandralis_minimum_2023} and LEONARDO \cite{kim_bipedal_2021,liang_rough-terrain_2021,sihite_optimization-free_2021,sihite_efficient_2022}. The M4 robot aimed to enhance locomotion versatility by combining posture manipulation and thrust vectoring, enabling walking, wheeling, flying, and loco-manipulation. LEONARDO, a quadcopter with two legs, can perform both quasi-static walking and flying.

However, neither of these robots fully demonstrates dynamic legged locomotion and aerial mobility. Integrating both modes remains a significant challenge due to conflicting requirements (see \cite{sihite_multi-modal_2023}). Achieving dynamic walking and aerial mobility within a single platform continues to be a major obstacle in hardware and control design.

The Harpy design is unique for two main reasons, distinguishing it from M4 and LEONARDO:

\begin{itemize}
    \item Harpy's thrusters cannot fully stabilize vehicle dynamics, meaning Harpy is not a quadcopter.
    \item Harpy's joint actuators support dynamic bipedal legged locomotion, making Harpy a bipedal robot.
\end{itemize}

\begin{figure*}[h!]
    \vspace{0.08in}
    \centering
    \includegraphics[width=\linewidth]{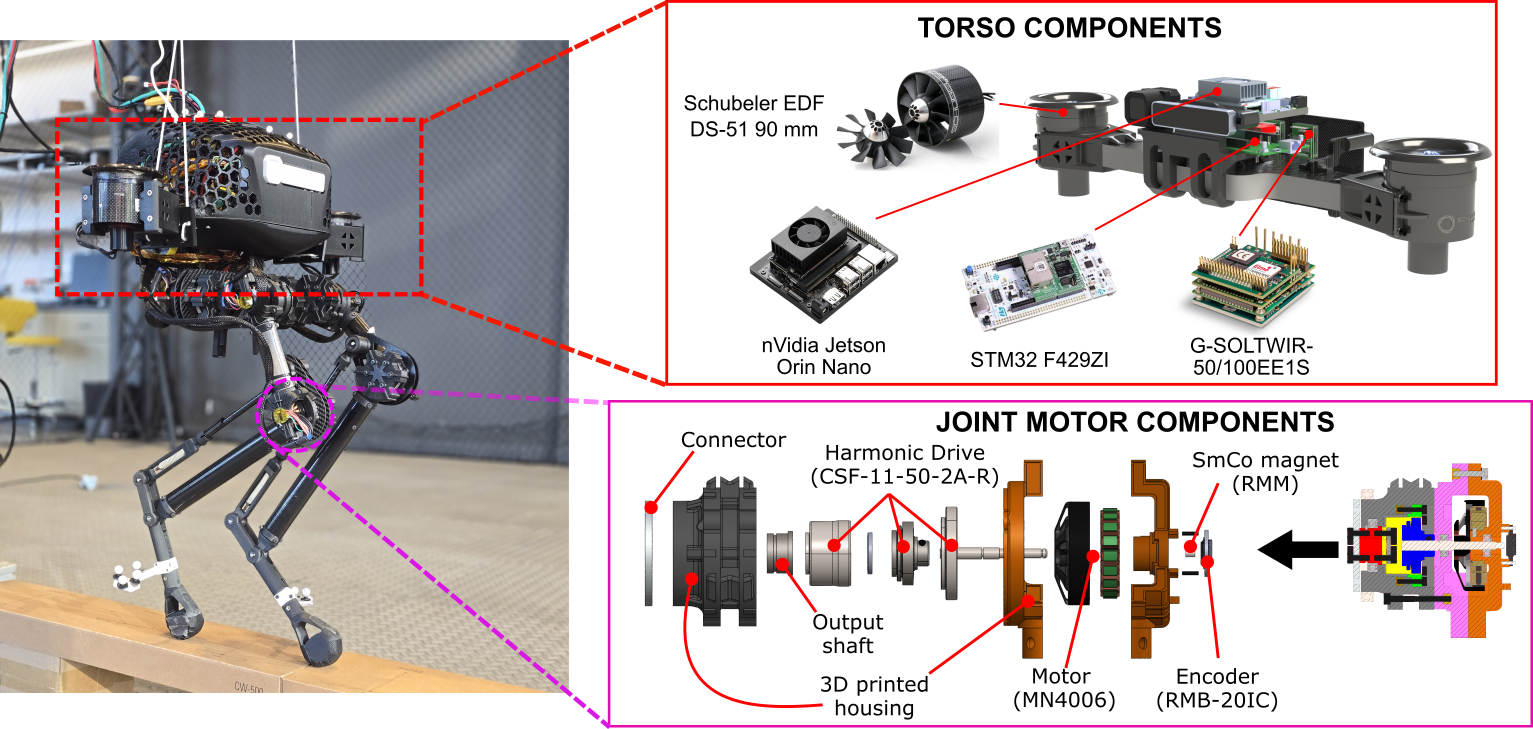}
    \caption{Illustration showing the Harpy robot with its rendered model, and its torso and joint motor components. The torso is built on a lightweight beam made out of an aluminum honeycomb structure sandwiched between thin carbon fiber plates. Core power and microcontroller components are placed within the torso. The custom joint actuator design utilizes a brushless motor and a Harmonic Drive for a small, compact, and lightweight motor-gearbox design that does not compromise its power.}
    \vspace{-0.08in}
    \label{fig:detailed-view}
\end{figure*}

This design offers unique control opportunities that have yet to be explored, such as high jumps and walking on steep or vertical surfaces. One such opportunity, and the focus of this paper, is the stabilization of frontal dynamics using thrusters.

In dynamic bipedal walkers, sagittal, frontal, and transversal dynamics are controlled primarily through foot placement. For transversal dynamics, humanoids require additional joints to allow leg rotation in the transversal plane. Here, not only do we have the means to stabilize frontal dynamics using foot placement, but we can also use thrusters to increase robustness against pushovers (though this is not demonstrated in this paper).

More specifically, we employ a detailed model of Harpy (a SimScape high-fidelity model, as shown in Fig.~\ref{fig:mdl-params}). We derived a reduced-order model of the system based on the variable-length inverted pendulum (VLIP) model. To stabilize the frontal dynamics, we implemented a high-derivative gain thruster controller, while for the sagittal dynamics, we used a capture point controller \cite{pitroda_dynamic_2023,dangol_feedback_2020-1,dangol_control_2021,pitroda_capture_2024}.

While capture point control based on centroidal models for bipedal systems has been extensively studied \cite{koolen_design_2016,ramos_generalizations_2015,hong_capture_2019,bickel_capture_2009}, the incorporation of thruster forces that influence the dynamics of linear inverted pendulum models—often used in capture point-based works—has not been explored extensively. The inclusion of these external thrust forces opens up novel interpretations of locomotion, such as the concept of virtual buoyancy, which has been explored in aquatic-legged locomotion \cite{iqbal_extended_2021,krause_stabilization_2012,pratt_capture_2006}.

{
This work is structured as follows: introduction and hardware overview, followed by modeling and control design, experimental results and discussion, and the concluding remarks.}

\section{Hardware Overview}
\label{sec:hdw}

Harpy (Fig.~\ref{eq:pos_com},~\ref{fig:detailed-view}) is a bipedal robot with two upward-facing thrusters attached to the side of its torso, as shown in Fig.\ref{fig:cover-image}. The robot weighs approximately 7 kg and each thruster can generate approximately 2.5 kg maximum thrust, resulting in a 1.6 thrust-to-weight ratio (agile qaud-copters possess a 2 ratio).  The thrusters are aligned as close to the body's center of mass (CoM) as possible to minimize thruster pitch moment. The robot has 12 degrees of freedom (DOF): three joints per leg (hip frontal, hip sagittal, and knee joints), body position and orientation.

The robot's body is primarily composed of lightweight materials such as carbon fiber tubes, plates, and reinforced 3D-printed components. The torso is built around a thick but lightweight beam made out of an aluminum honeycomb structure sandwiched between two thin carbon fiber plates. The torso housed all of the important electronics such as the microcontrollers, thrusters, and motor drivers. The legs are composed of oval carbon fiber tubes connected through 3D-printed structures. The 3D printed structures were printed using a MarkForged 3D printer with added carbon fiber reinforcing material, which increases the part's stiffness and strength. The robot's components are shown in more detail in Fig.~\ref{fig:detailed-view}.

The joint motors are custom-made using T-Motor and Harmonic Drive (50:1 gear ratio for the knee motors and 30:1 for the rest), and they are driven by ELMO Gold Twitter where the motor commands are sent through EtherCAT communication protocol. Hall effect encoders are utilized at the joints to measure joint angles and track the desired joint positions. The thrusters are Schubeler DS-51 90mm Electric Ducted Fans (EDF), which can provide 5kg maximum thrust at 50V reference voltage.  The propeller motors are powered using APD 120F3[X] ESC, rated up to 12S (50.4V) and 120A continuous current. The system is powered using two 6S batteries connected in series, resulting in a reference voltage of 50V when the batteries are fully charged.

The robot is controlled using Simulink Realtime through etherCAT communication protocol between the xPC Target computer, ELMO amplifiers, and the Nucleo STM32 F429ZI board. The STM32 F429ZI board is attached to a Hilscher netX board which enables communication between EtherCAT and non-EtherCAT components, primarily the ESCs which must be driven using a servo PWM signal and motion capture data which is processed through serial communication. The STM32 F429ZI board also communicates to the Jetson Orin Nano, which is a powerful computer that can be used for computationally expensive algorithms such as MPC and localization. The controller loop runs at 500 Hz for fast joint tracking while the motion while the motion capture can deliver pose data at 240 Hz.

\section{Electronics and Prototyping}
\label{chap:harpy:electronics}
The Harpy Robot employs a sophisticated multi-layered communication architecture for precise motion control and state estimation. At the highest level, eight OptiTrack cameras operate at 360 Hz, capturing infrared reflections from markers mounted on the robot's head. These cameras connect via Power over Ethernet (PoE+) cables, delivering both power (30W per port) and Gigabit Ethernet data transmission.

The motion capture data flows from the OptiTrack computer to a Linux machine using the NatNet protocol. This protocol ensures efficient streaming of 6-DoF pose data, including position vectors and quaternion orientation, with system latency of approximately 2.8 ms.

\begin{figure}[h]
  \centering
    \includegraphics[width=\textwidth]{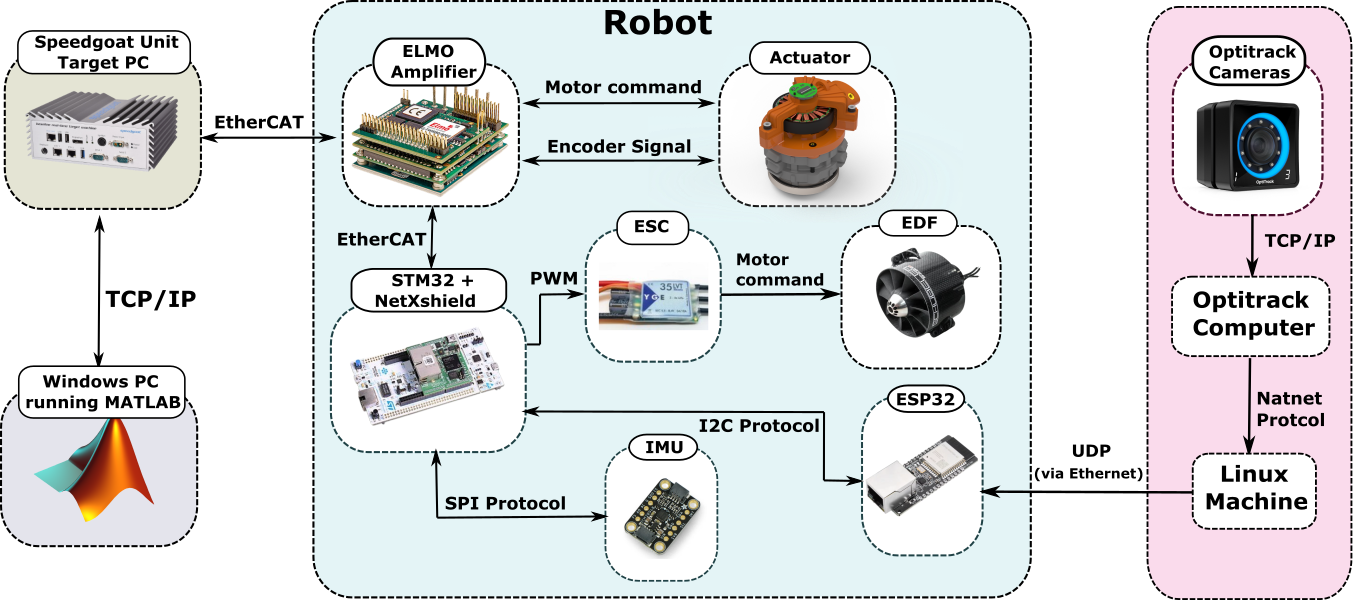}
  \caption[Harpy Architecture]{Harpy Architecture}
  \label{fig:harpy_arch}
\end{figure}

The Linux computer processes this pose information within the ROS framework and transmits control commands to an ESP32 microcontroller via Ethernet using UDP protocol at 300Hz. The UDP implementation provides low-latency communication without connection overhead, utilizing a simple 8-byte header structure for efficient data transfer.

The ESP32 communicates with an STM32 F429ZI microcontroller through I2C protocol. Initially, the system experienced communication issues due to static charges, which was resolved by implementing 5k$\Omega$ pull-up resistors on both SDA and SCL lines to ensure reliable data transmission.

The STM32 processes both position data from the ESP32 and orientation data from the IMU, consolidating this information for transmission over the EtherCAT network to the target computer. This ensures synchronized control and state feedback for the robot's six actuated joints. For thrust control, the STM32 generates PWM signals for the ESCs controlling the left and right thrusters. This completes the control chain from high-level motion capture to low-level actuator commands, enabling coordinated control of both joint movements and thrust generation.

An ICM-20948 9-axis IMU connects to the STM32 via SPI protocol, providing orientation data at 200 Hz. The system encountered EMI noise issues due to proximity to Elmo motor drivers, which was mitigated through copper tape shielding. The IMU operates with configurable ranges and digital filtering options to optimize performance.

The STM32 interfaces with a Hilscher netX netSHIELD board via SPI protocol, enabling EtherCAT communication. This board features dual RJ45 ports supporting line and ring topologies, facilitating real-time industrial ethernet connectivity.

The EtherCAT network connects to six Elmo amplifiers along with the STM32 NetX Shield in a daisy-chain configuration, with the entire network terminating at a Speedgoat real-time target computer. The Elmo amplifiers support advanced motion control features and integrate seamlessly with the EtherCAT protocol. The control architecture runs on a Simulink Real-Time system, which enables rapid prototyping and hardware-in-the-loop testing. This platform provides deterministic performance for complex control algorithms and supports various communication interfaces including EtherCAT.
\\

\section{Modeling}
\label{sec:model}

This section outlines the dynamics formulation of the robot used for simulation and control design (Section~\ref{sec:ctrl}). The left-hand side of Fig.~\ref{fig:mdl-params} shows the high-fidelity model parameters, including joint positions, CoM locations, and inputs (actuation torques and thrusters).

\subsection{Energy-based Lagrange Formalism}

\begin{figure}[h!]
    \vspace{0.08in}
    \centering
    \includegraphics[width=\linewidth]{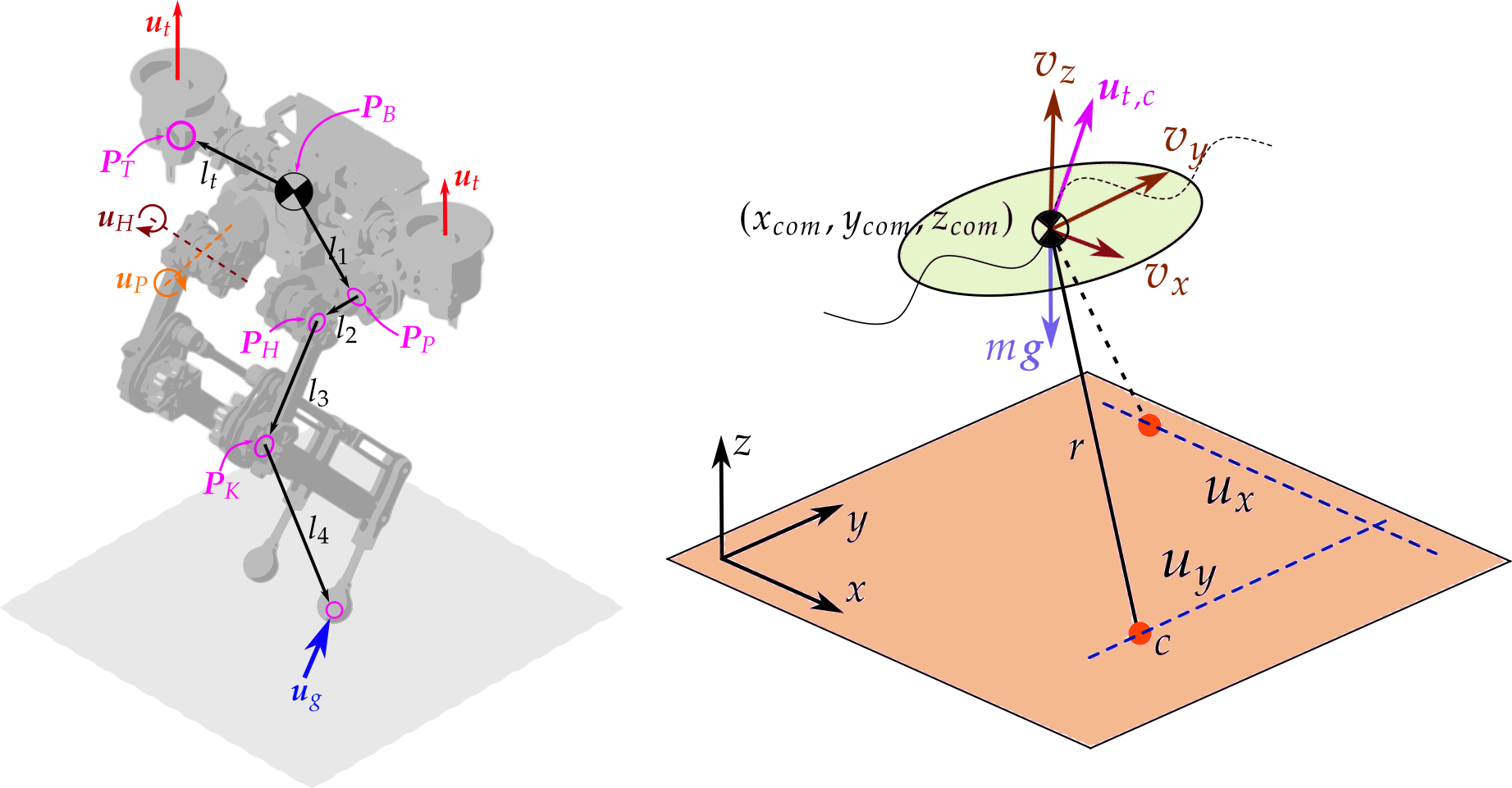}
    \caption{The left image shows the parameters of our high-fidelity robot, while the right image shows the reduced-order model used for control design.}
    \vspace{-0.08in}
    \label{fig:mdl-params}
\end{figure}

The Harpy equations of motion are derived using Euler-Lagrangian dynamics formulation. In order to simplify the system, each linkage is assumed to be massless, with the mass concentrated at the body and the joint motors. Consequently, the lower leg kinematic chain is considered massless, significantly simplifying the system. The three leg joints are labeled as the hip frontal (pelvis $P$), hip sagittal (hip $H$), and knee sagittal (knee $K$), as illustrated in Fig. \ref{fig:mdl-params}. 

Let $\gamma_h$ be the frontal hip angle, while $\phi_h$ and $\phi_k$ represent the sagittal hip and knee angles, respectively. The superscripts $\{B,P,H,K\}$ represent the frame of reference about the body, pelvis, hip, and knee, while the inertial frame is represented without the superscript. Let $R_B$ be the rotation matrix from the body frame to the inertial frame (i.e., $\bm x = R_B\, \bm x^B$). The pelvis motor mass is added to the body mass. Then, the positions of the hip and knee centers of mass (CoM) are defined using kinematic equations:
\begin{equation}
\begin{gathered}
    \bm{p}_P = \bm{p}_{B} + R_{B}\, \bm{l}_{1}^{B}, \\
    \bm{p}_H = \bm{p}_{P} + R_{B}\,R_x(\gamma_h)\, \bm{l}_{2}^{P} \\
    \bm{p}_K = \bm{p}_{H} + R_{B}\,R_x(\gamma_h)\,R_y(\phi_h) \bm{l}_{3}^{H},
\end{gathered}
\label{eq:pos_com}
\end{equation}
where $R_x$ and $R_y$ are the rotation matrices about the $x$ and $y$ axes, respectively, and $\bm l$ is the length vector representing the configuration of Harpy, which remains constant in its respective local frame of reference. The positions of the foot and thrusters are defined as:
\begin{equation}
\begin{gathered}
    \bm{p}_F = \bm{p}_{K} + R_{B}\,R_x(\gamma_h)\,R_y(\phi_h)\,R_y(\phi_k)\, \bm{l}_{4}^{K} \\
    \bm{p}_T = \bm{p}_{B} + R_{B}\, \bm{l}_{t}^{B}
\end{gathered}
\label{eq:pos_other}
\end{equation}
where the length vector from the knee to the foot is $\bm l_4^K = [-l_{4a}\cos{\phi_k}, 0, -( l_{4b} + l_{4a}\sin{\phi_k})]^\top$, which represents the kinematic solution to the parallel linkage mechanism of the lower leg. Let $\bm \omega_B$ be the angular velocity of the body. Then, the angular velocities of the hip and knee are defined as: $\bm \omega_H^B = [\dot{\gamma}_h,0,0]^\top + \bm \omega_B^B$ and $\bm \omega_K^H = [0,\dot{\phi}_h,0]^\top + \bm \omega_H^H$. Consequently, the total energy of Harpy for the Lagrangian dynamics formulation is defined as follows:
\begin{equation}
\begin{aligned}
    K &= \tfrac{1}{2} \textstyle \sum_{i \in \mathcal{F}} \left( 
        m_i\,\bm p_i^\top\, \bm p_i + 
        \bm \omega_i^{i \top} \, \hat I_i \, \bm \omega_i^i \right) \\
    V &= - \textstyle \sum_{i \in \mathcal{F}} \left( 
        m_i\,\bm p_i^\top\, [0,0,-g]^\top \right),
\end{aligned}
\label{eq:energy}
\end{equation}
where $\mathcal{F} = \{B,H_L,K_L,H_R,K_R\}$ represents the relevant frames of reference and mass components (body, left hip, left knee, right hip, right knee), and the subscripts $L$ and $R$ denote the left and right sides of the robot, respectively. Furthermore, $\hat I_i$ denotes the inertia about its local frame, and $g$ is the gravitational constant. This constitutes the Lagrangian of the system, given by $L = K - V$, which is utilized to derive the Euler-Lagrange equations of motion. 

The dynamics of the body's angular velocity are derived using the modified Lagrangian for rotation in $SO(3)$ to avoid using Euler angles and the potential gimbal lock associated with them. This yields the following equations of motion following Hamilton's principle of least action:
\begin{equation}
\begin{gathered}
    \tfrac{d}{dt}\left( \tfrac{\partial L}{\partial \bm \omega_B^B}  \right) + 
    \bm \omega_B^B \times \tfrac{\partial L}{\partial \bm \omega_B^B} + 
    \textstyle \sum_{j=1}^{3} \bm r_{Bj} \times \tfrac{\partial L}{\partial \bm r_{Bj}} = \bm u_1, \\
    \tfrac{d}{dt}\left( \tfrac{\partial L}{\partial \dot {\bm q}}  \right) - 
    \tfrac{\partial L}{\partial \bm q} = \bm u_2, \\ 
    \tfrac{d}{dt} R_B = R_B\, [\bm \omega_B^B]_\times,
\end{gathered}
\label{eq:eom_eulerlagrange}
\end{equation}
where $[\, \cdot \, ]_\times$ denotes the skew symmetric matrix, $R_B^\top = [\bm r_{B1}, \bm r_{B2}, \bm r_{B3}]$, $\bm q = [\bm p_B^\top, \gamma_{h_L}, \gamma_{h_R}, \phi_{h_L}, \phi_{h_R}]^\top$ represents the dynamical system states other than $(R_B,\bm \omega^B_B)$, and $\bm u$ denotes the generalized forces. The knee sagittal angle $\phi_k$, which is not associated with any mass, is updated using the knee joint acceleration input $\bm u_k = [\ddot{\phi}_{k_L}, \ddot{\phi}_{k_R}]^\top$. Then, the system acceleration can be derived as follows:
\begin{equation}
\begin{gathered}
    M \bm a + \bm h = B_j\, \bm u_j + B_t\, \bm u_t + B_g\, \bm u_g
\end{gathered}
\label{eq:eom_accel}
\end{equation}
where $\bm a = [ \dot{\bm \omega}_B^{B\top}, \ddot{\bm q}^\top, \ddot{\phi}_{k_L}, \ddot{\phi}_{k_R}]^\top$, $\bm u_t$ denotes the thruster force, $\bm u_j = [u_{P_L}, u_{P_R}, u_{H_L}, u_{H_R}, \bm u_k^\top]^\top$ represents the joint actuation, and $\bm u_g$ stands for the ground reaction forces (GRFs). The variables $M$, $\bm h$, $B_t$, and $B_g$ are functions of the full system states:
\begin{equation}
    \bm x = [\bm r_{B}^\top, \bm q^\top, \phi_{K_L}, \phi_{K_R}, \bm \omega_B^{B \top}, \dot{\bm q}^\top, \dot{\phi}_{K_L}, \dot{\phi}_{K_R}]^\top,
\label{eq:states}
\end{equation}
where the vector $\bm r_B$ contains the elements of $R_B$. Introducing $B_j = [0_{6 \times 6}, I_{6 \times 6}]$ allows $\bm u_j$ to actuate the joint angles directly. Let $\bm v = [\bm \omega_B^{B\top}, \dot{\bm q}^\top]^\top$ denote the velocity of the generalized coordinates. Then, $B_t$ and $B_g$ can be defined using the virtual displacement from the velocity as follows:
\begin{equation}
\begin{aligned}
    B_t = \begin{bmatrix}
        \begin{pmatrix}
        \partial \dot{\bm p}_{T_L} / \partial \bm v \\
        \partial \dot{\bm p}_{T_R} / \partial \bm v
        \end{pmatrix}^\top
        \\
        0_{2 \times 6}
    \end{bmatrix}, \quad
    B_g = \begin{bmatrix}
        \begin{pmatrix}
        \partial \dot{\bm p}_{F_L} / \partial \bm v \\
        \partial \dot{\bm p}_{F_R} / \partial \bm v
        \end{pmatrix}^\top
        \\
        0_{2 \times 6}
    \end{bmatrix}.
\end{aligned}
\label{eq:generalized_forces}
\end{equation}
The vector $\bm u_t = [\bm u_{t_L}^\top, \bm u_{t_R}^\top]^\top$ is composed of the left and right thruster forces $\bm u_{t_L}$ and $\bm u_{t_R}$, respectively. The GRF is modeled using the unilateral compliant ground model with undamped rebound \cite{liang_rough-terrain_2021}.

\subsection{Raibert Foot Placement Model}

The Raibert controller is based on a simple yet effective principle that relates the Center of Mass (CoM) dynamics to foot placement strategy. The core concept involves placing the foot at a position that will help maintain the desired forward velocity and balance.

The fundamental equations of the Raibert controller are given by
\begin{equation}
\begin{aligned}
u_x &= \alpha_x v_x + \beta_x + x_{com} \\
u_y &= \alpha_y v_y + \text{sign}(y) \beta_y + y_{com}
\end{aligned}
\end{equation}
where $\alpha_x$, $\alpha_y$ are velocity feedback gains, $\beta_x$, $\beta_y$ are position offset terms, and $x_{com}$, $y_{com}$ represent the CoM position relative to the stance foot.

The robot's complete state is captured by
\begin{equation}
\mathbf{x} = \begin{bmatrix} 
p_{com} \\
v_{com} \\
q \\
\omega
\end{bmatrix}
\end{equation}
which includes the CoM position $p_{com}$, velocity $v_{com}$, orientation quaternion $q$, and angular velocity $\omega$.

\section{Controls}
\label{sec:ctrl}

\subsection{Sagittal Dynamics Stabilization with Foot Placement}
\label{sec:ctrl:sagittal}
The controller uses phase variables to track the progress through the gait cycle:
\begin{equation}
\begin{aligned}
s_D &= \frac{t - t_{start}}{T_D}, \quad s_D \in [0,1] \\
s_S &= \frac{t - t_D}{T_S}, \quad s_S \in [0,1]
\end{aligned}
\end{equation}
Here, $s_D$ represents the double support phase progress, normalized from 0 to 1, where $t_{start}$ is the phase start time and $T_D$ is the total double support duration. Similarly, $s_S$ tracks single support phase progress over duration $T_S$.

The controller handles three distinct cases during locomotion:

Case I: Initial Single Support ($s_D = 1$ and $s_S = 0$)
For Left Leg (state = 1): \\
\begin{equation}
\begin{aligned}
p_{target} &= p_{inter} \\
p_{apex\_des} &= \begin{bmatrix}
p_{apex,x} & p_{apex,y} & p_{apex,z} & p_{apex,x} & 0 & 0
\end{bmatrix}^T
\end{aligned}
\end{equation}
During left leg support, $p_{target}$ maintains the intermediate position while $p_{apex\_des}$ defines the desired apex (midpoint) position for the swing trajectory. The zeros in the lower half indicate no movement for the stance leg.

For Right Leg  (state = -1): \\
\begin{equation}
\begin{aligned}
p_{target} &= p_{inter} \\
p_{apex\_des} &= \begin{bmatrix}
p_{apex,x} & 0  &0  &p_{apex,x}  &p_{apex,y}& p_{apex,z}
\end{bmatrix}^T
\end{aligned}
\end{equation}
Similarly for right leg support, but with the apex positions defined for the opposite leg.

Case II: Double Support ($s_D < 1$ and $s_S = 0$)
\begin{equation}
\begin{aligned}
p_{target} &= p_{inter} + \begin{cases}
\begin{bmatrix}0 & \delta y_L & 0 & 0 & \delta y_L & 0\end{bmatrix}^T
 & \text{if } state = 1 \\
\begin{bmatrix}0 & -\delta y_R & 0 & 0 & -\delta y_R & 0\end{bmatrix}^T & \text{if } state = -1
\end{cases} \\
p_{apex\_des} &= \mathbf{0}_{6 \times 1}
\end{aligned}
\end{equation}
During double support, lateral adjustments ($\delta y_L$ and $\delta y_R$) are added to the intermediate position. The apex desired position is zeroed as no swing motion is needed.

The intermediate positions are calculated using:
\begin{equation}
p_{inter} = p_{init} + p_{\delta} + \begin{bmatrix}
\delta pf_{des,x} & -\delta pf_{des,y} & \delta pf_{des,z} &
\delta pf_{des,x} & \delta pf_{des,y} & \delta pf_{des,z}
\end{bmatrix}^T
\end{equation}
where $p_{init}$ is the neutral stance position, $p_{\delta}$ adds position offsets, and the last term adds desired adjustments in each direction. Note the sign change in $\delta pf_{des,y}$ for left/right symmetry.

Case III: Raibert Single Support ($s_D = 1$ and $s_S > 0$)
For Left Swing (state = 1):
\begin{equation}
\begin{aligned}
p_{target} &= \begin{bmatrix}
\delta pos_{swing,x} + p_{offset,x} &
\delta pos_{swing,y} + p_{offset,y} & \\
p_{foot\_target,z} & 
\delta pos_{stance,x} + p_{offset,x'} & \\
\delta pos_{stance,y} + p_{offset,y'} &
p_{foot\_target,z'} \cdot \delta  pos_{stance,z}
\end{bmatrix}^T
\end{aligned}
\end{equation}
During left leg swing, the target position combines:
- Swing leg adjustments ($\delta pos_{swing}$)
- Temporary position offsets ($p_{offset}$)
- Desired foot height ($p_{foot\_target,z}$)
- Stance leg adjustments with scaling for height

For Right Swing (state = -1): \\
\begin{equation}
\begin{aligned}
p_{target} &= \begin{bmatrix}
\delta pos_{stance,x} + p_{offset,x} &
\delta pos_{stance,y} + p_{offset,y} & \\
p_{foot\_target,z} \cdot \delta pos_{stance,z} &
\delta pos_{swing,x} + p_{offset,x'} & \\
\delta pos_{swing,y} + p_{offset,y'} &
p_{foot\_target,z'}
\end{bmatrix}^T
\end{aligned}
\end{equation}
The right swing case mirrors the left swing, with stance and swing components swapped.

The smooth transition between positions is achieved using a Bezier curve:
\begin{equation}
B(s) = \sum_{i=0}^n \begin{pmatrix} n \\ i \end{pmatrix} (1-s)^{n-i} s^i P_i
\label{eqn:bezier}
\end{equation}
where:
- $n$ is typically 5 or 6 for smooth motion
- $P_i$ are control points defining the path
- $s$ is the normalized time parameter (0 to 1)
- The binomial coefficient and power terms ensure smooth blending between control points

\subsection{Frontal Dynamics stabilization with Thrusters}
\label{subsec:frontal_control}

The thrusters attached to the robot can generate high-response-time roll moments (a characteristic of ducted fans as opposed to exposed propellers) to stabilize the frontal dynamics. As a result, frontal stabilization is achieved using a PID control through differential thrust control. The system state is represented by the orientation angles and their derivatives:
\begin{equation}
\mathbf{q} = \begin{bmatrix} 
\phi \\
\theta \\
\dot{\phi} \\
\dot{\theta}
\end{bmatrix}
\end{equation}
$\phi$ represents the roll angle about the x-axis, $\theta$ is the pitch angle about the y-axis, and their respective angular velocities $\dot{\phi}$ and $\dot{\theta}$ complete the state vector.

The control error is computed as the difference between the reference state and current state:
\begin{equation}
\mathbf{e_\phi} = \mathbf{q}_{ref} - \mathbf{q}
\end{equation}
$\mathbf{q}_{ref}$ contains the desired orientation and angular velocities.

The integral error term is computed with anti-windup protection:
\begin{equation}
e_{\phi,i} = \text{sat}(\int_0^t e_\phi(t) dt , \pm\frac{\text{max}_i}{K_i})
\end{equation}
$\text{max}_i$ limits the maximum integral contribution and $K_i$ is the integral gain. The saturation function prevents integral windup by clamping the accumulated error.

The control law combines proportional, integral, and derivative terms:
\begin{equation}
u_t = K_p e_\phi + K_i e_{\phi,i} + K_d \dot{e}_\phi
\end{equation}
where:
- $K_p$ provides proportional correction to current error
- $K_i e_{\phi,i}$ adds integral action to eliminate steady-state error
- $K_d \dot{e}_\phi$ adds derivative damping for stability

The final thrust commands for left and right thrusters are computed as:
\begin{equation}
\begin{bmatrix} 
u_L \\
u_R
\end{bmatrix} = \text{sat}\begin{bmatrix}
u_{base} + u_t \\
u_{base} - u_t
\end{bmatrix}_{[u_{min}, u_{max}]}
\end{equation}
where:
- $u_{base}$ is the nominal thrust for hovering
- $u_t$ is added/subtracted differentially to create corrective moments
- The saturation function ensures commands stay within actuator limits $[u_{min}, u_{max}]$
- $u_L$ and $u_R$ are the final thrust commands for left and right thrusters

\begin{figure}[h!]
    \vspace{0.08in}
    \centering
    \includegraphics[width=0.5\linewidth]{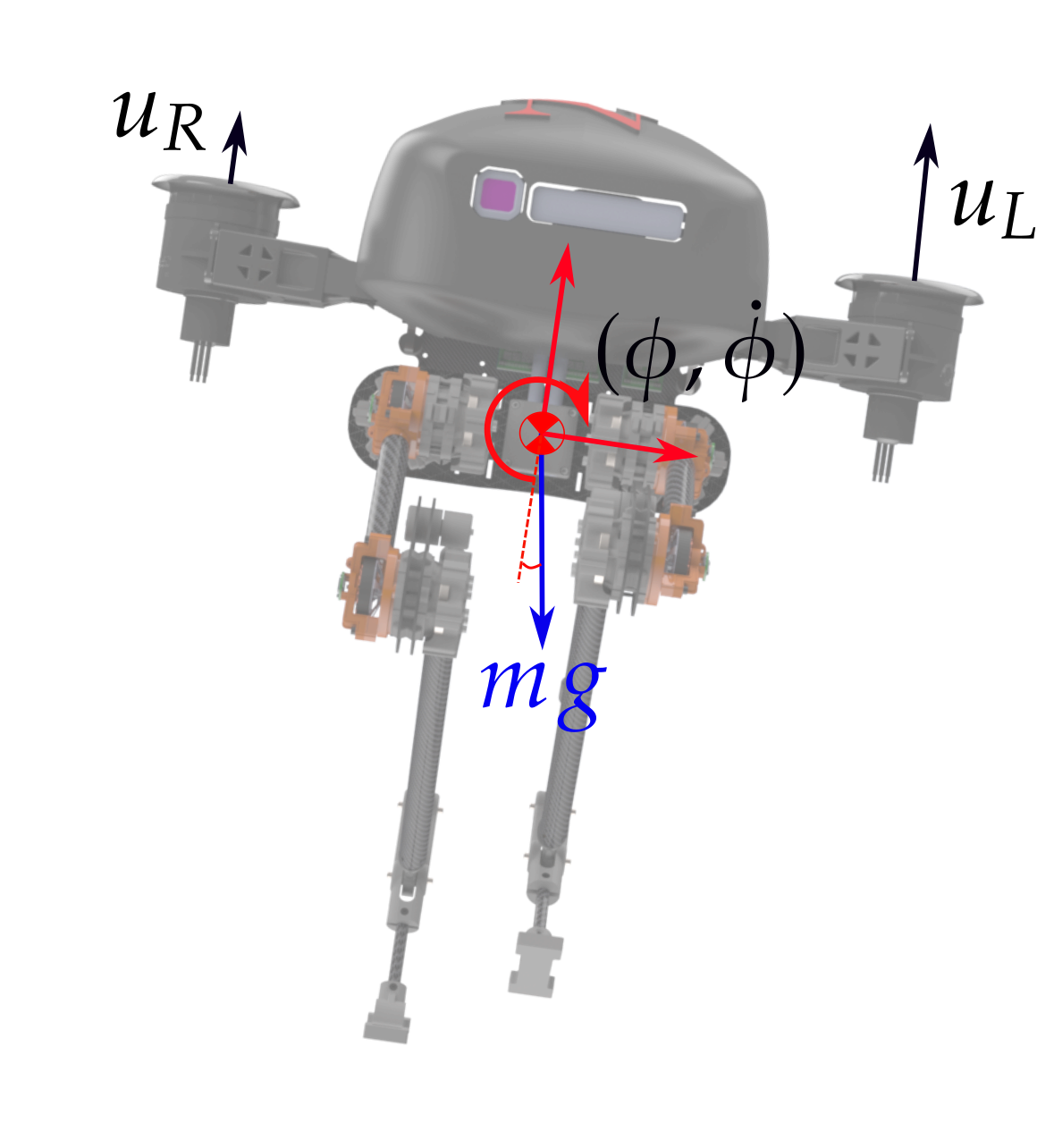}  
    \caption{FBD of Flight Controller}
    \vspace{-0.08in}
    \label{fig:flight_controller}
\end{figure}

This control structure provides stable orientation control while handling actuator constraints and preventing integral windup effects.

\subsection{Control Architecture}

\begin{figure}[t]
    \vspace{0.08in}
    \centering
    \includegraphics[width=0.9\linewidth]{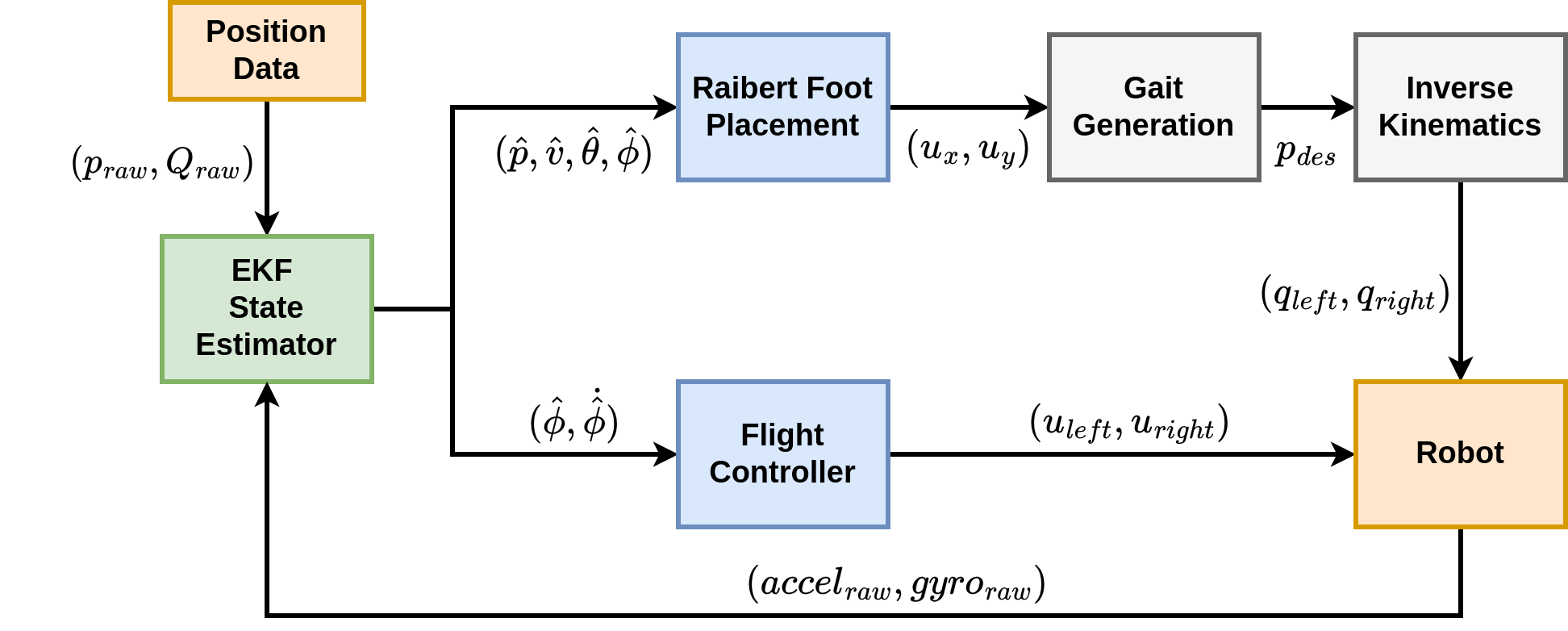}  
    \caption{Control Architecture}
    \vspace{-0.08in}
    \label{fig:control_arch}
\end{figure}

This is the combined control architecture for Harpy robot, it consist of the following:

\begin{enumerate}
    \item Position Data: This data is the pose information from the robot COM, which is collected using the OptiTrack Motion Capture system. It is in the form of position coordiantes $p_{raw} = (x, y, z)$ and Quaternions $Q_{raw} = (q_1, q_2, q_3, q_4)$ and is recieved at 300Hz.
    \item Robot: The robot block represents Harpy, which receives target joint angles from inverse kinematics, thrust values from flight controller and gives output of raw accelerometer and gyroscope values throught the IMU at 200Hz.
    \item EKF State Estimator : It outputs the estimated position state of the robot at 500Hz  by sensor fusion of optitrack position data $(p_{raw}, Q_{raw})$ and raw IMU measurements $(accel_{raw}, gyro_{raw})$. The estimated pose information is given out to both the controllers: Raibert Foot Placement and Flight Controller.
    \item Raibert Foot Placement: This controller is responsible to stabilize sagittal plane of robot and output foot placement distances. It recieves the input as the estimated states from EKF estimator which are: $(\hat{p}, \hat{v}, \hat{\theta}, \hat{\phi})$, and the outputs are the foot placement distance in x and y axis: $(u_x, u_y)$ the controller is described in more detailed at Section ~\ref{sec:ctrl:sagittal}.
    \item Flight Controller: This controller is responsible to stabilize the frontal dynamics of the robot where it calculates the output thrust for both the thrusters. Its input are the roll angle and rate of change of roll angle from the estimator: $(\hat{\phi}, \dot{\hat{\phi}})$. The controller outputs the left and right thrust $(u_{left}, u_{right})$. The controller architecture is described in Section ~\ref{subsec:frontal_control}.
    \item Gait Generation: The gait generation computes the intermediate foot end locations using bezier curve. Its denoted by Eqn. ~\ref{eqn:bezier}. The inputs are the updated foot locations from Raibert controller and it computes the desired foot trajectories $p_{des}$ for that input distance.
    \item Inverse Kinematics: The inverse Kinematics computes the desired joint angle of all the six actuator joints using the Inverse Kinematics of the robot which is predefined using the distances of legs. It sends the computed joint angle to reach the input foot end distance.
\end{enumerate}

\section{Results}
\label{sec:res}

\begin{figure}[t]
    \vspace{0.08in}
    \centering
    \includegraphics[width=0.7\linewidth]{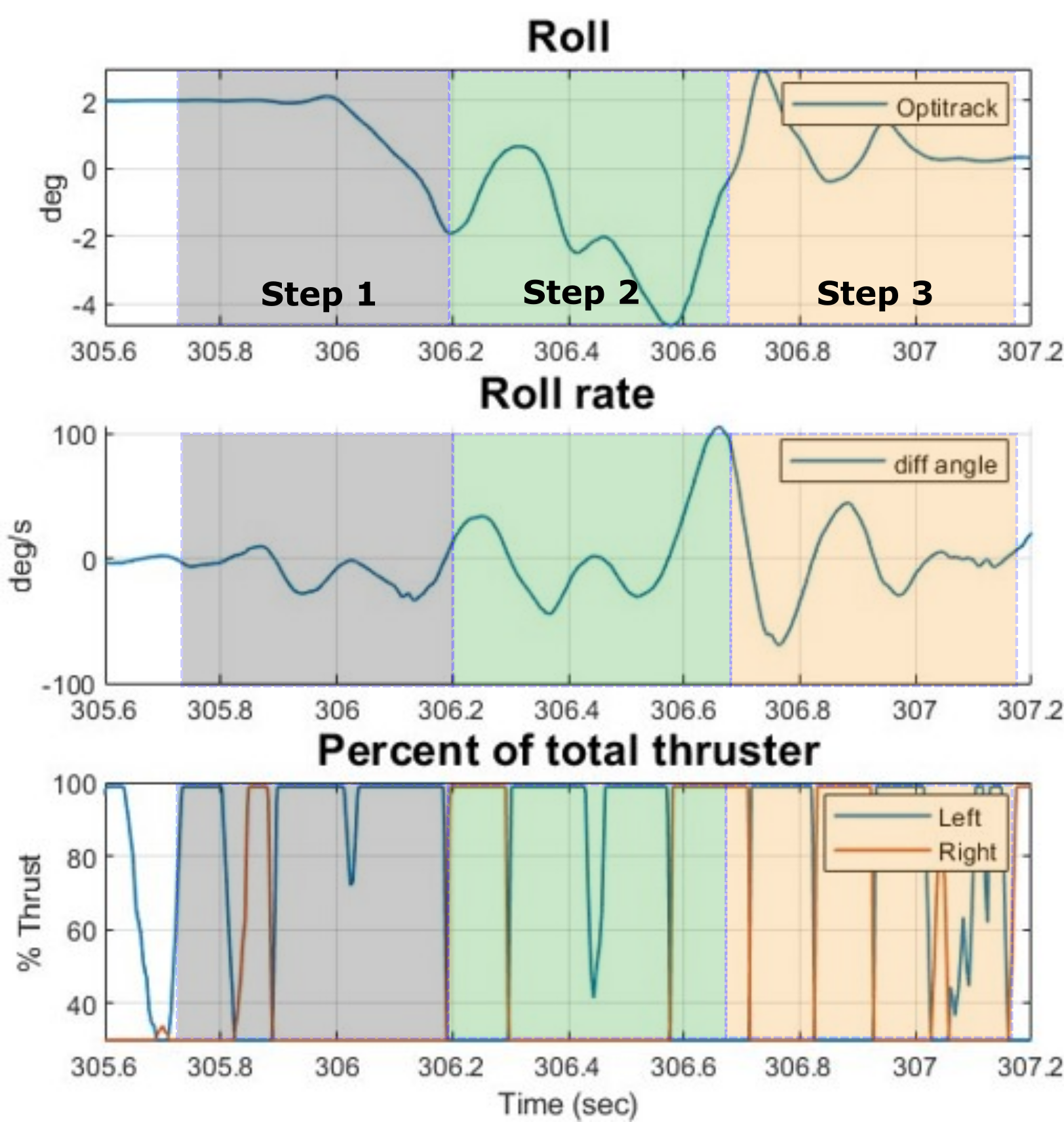}  
    \caption{Shows the Thruster Percentage wrt Roll angle}
    \vspace{-0.08in}
    \label{fig:thrust_percentage}
\end{figure}

\begin{figure}[h!]
    \vspace{0.08in}
    \centering
    \includegraphics[width=\linewidth]{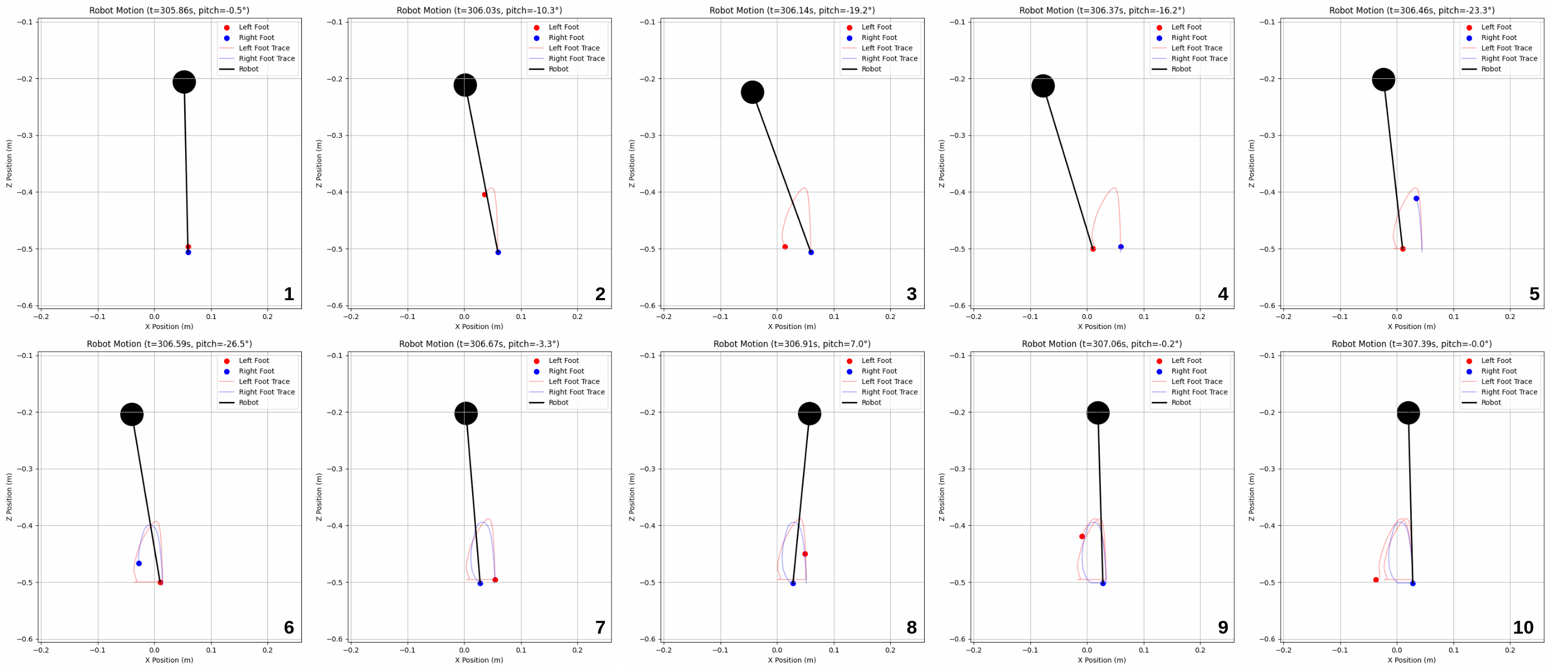}  
    \caption{Raibert Foot Placement in XZ Axis}
    \vspace{-0.08in}
    \label{fig:raibert_anim}
\end{figure}

\begin{figure}[h!]
    \vspace{0.08in}
    \centering
    \includegraphics[width=0.8\linewidth]{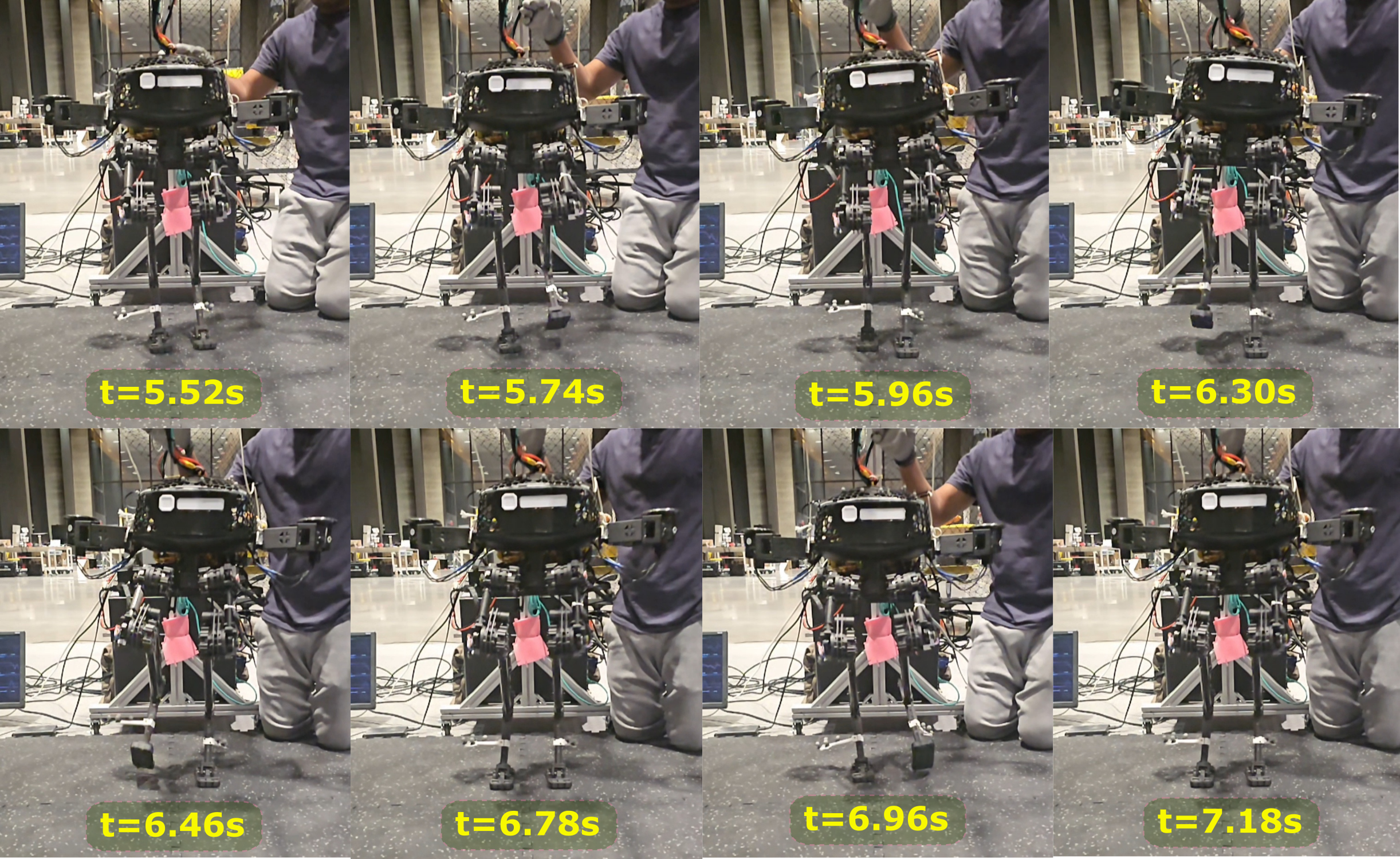} 
    \caption{Time-stamped Trotting Snapshots}
    \vspace{-0.08in}
    \label{fig:trotting_timestamp}
\end{figure}

\begin{figure*}
\vspace{0.08in}
    \centering
    \includegraphics[width=0.48\linewidth]{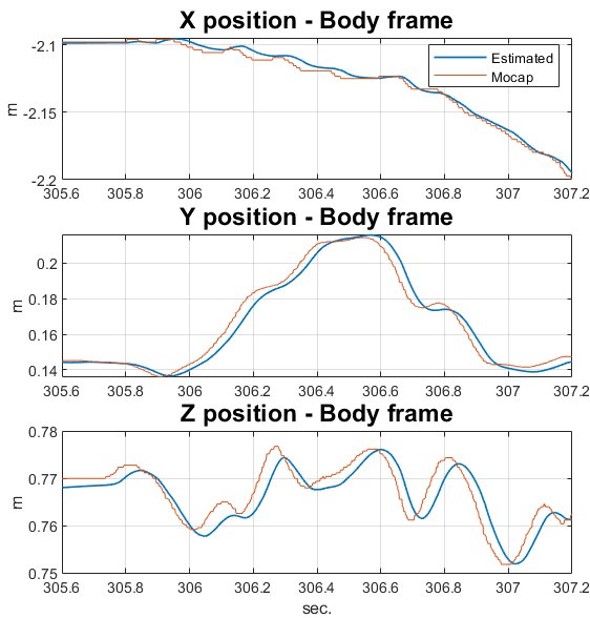}
    \hfill
    \includegraphics[width=0.48\linewidth]{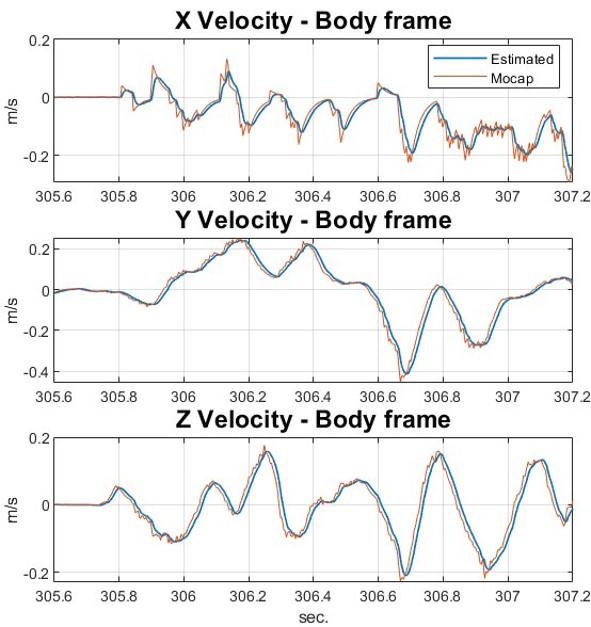}
    \caption{Shows the position and velocity of the COM from Optitrack vs estimated positions from EKF}
    \label{fig:body_pos_vel}
\vspace{-0.08in}
\end{figure*}

\begin{figure*}
\vspace{0.08in}
    \centering
    \includegraphics[width=0.48\linewidth]{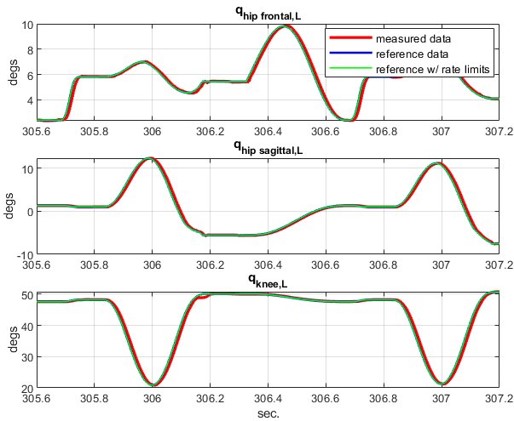}
    \hfill
    \includegraphics[width=0.48\linewidth]{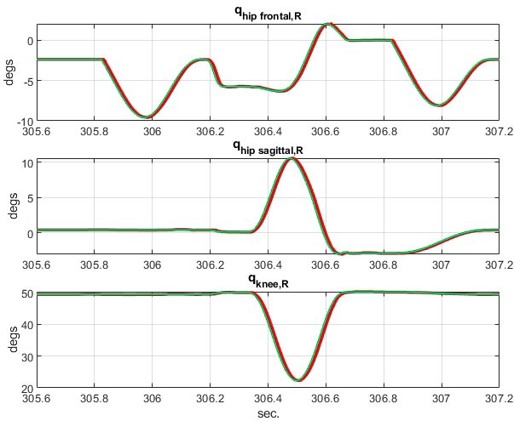}
    \caption{All 6 joint angles from both feet}
    \label{fig:joint_angles}
\vspace{-0.08in}
\end{figure*}

\begin{figure*}
\vspace{0.08in}
    \centering
    \includegraphics[width=0.49\linewidth]{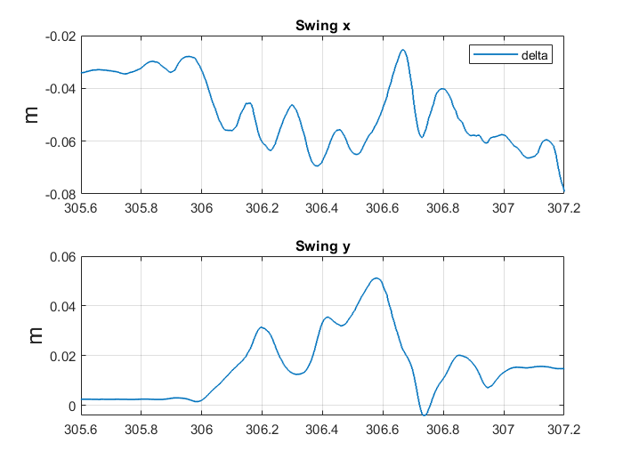}
    \hfill
    \includegraphics[width=0.49\linewidth]{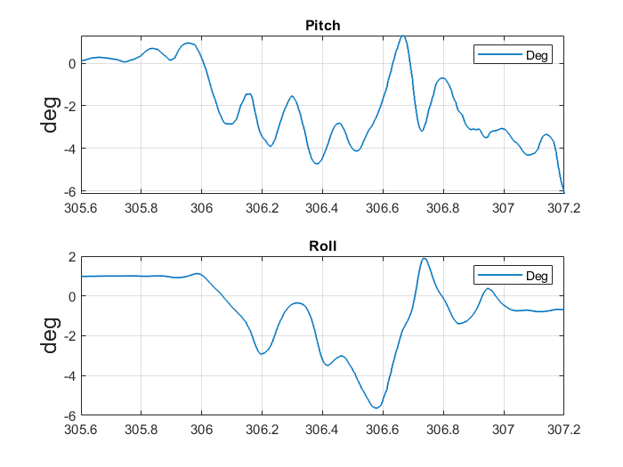}
    \caption{Foot swings from Raibert along EKF angles}
    \label{fig:raibert_ekf}
\vspace{-0.08in}
\end{figure*}



We performed a thruster-assisted trotting gait experiment to show that the thrusters can be used to stabilize the frontal dynamics. In this experiment, we utilized a motion capture system to estimate the robot's position and velocity in the inertial frame, in addition to the orientation using the sensor fusion between the motion capture attitude measurements and an on-board IMU. In this experiment, the robot performed a trotting gait in place and the frontal dynamics was stabilized using the derivative controller as described in Section~\ref{subsec:frontal_control}. 

The experimental results can be seen in Figures ~\ref{fig:trotting_timestamp} and ~\ref{fig:thrust_percentage}, where Fig.~\ref{fig:trotting_timestamp} shows the time-lapsed image of the robot performing the trotting gait and Fig.~\ref{fig:thrust_percentage} shows the measurements and control commands taken during the experiment. Figure ~\ref{fig:thrust_percentage} shows that the roll and pitch angles are stable throughout the experiment, and the stability of the roll angle shows that the thruster-assisted walking has successfully stabilized the robot's frontal dynamics. All of the states other than inertial x and y positions are relatively stable, as we do not have any position tracking controller active on the robot yet. Note that only one of the thruster base command value is set at 10\% to avoid any delays while generating thrust. This shows that the robot can walk without the thrusters actively generating lift to reduce the load on the leg joints. A position data collected using the EKF Estimation is done, the comparison between the optitrack and estimated position and velocity data can be seen in Fig.~\ref{fig:body_pos_vel}.

\section{Concluding Remarks}

In this work, we presented the mechanical design and preliminary experiments of our legged-aerial multi-modal robot, Harpy. In this experiment, we have demonstrated Harpy's ability to perform thruster-assisted dynamic walking where it can trot in place using minimal thruster assistance in frontal dynamic stabilization. This shows that the robot has enough leg joint power to walk without thrusters and can utilize its multi-modal capability to assist dynamic walking performance.

For future work, we will further explore the thruster-assisted walking ability of Harpy in scenarios where it is difficult or impossible to perform without thrusters, such as steep incline walking and narrow path walking. We will also explore Harpy's aerial mode capabilities by performing maneuvers and locomotion that are only possible using both legs and thrusters, such as high jump, soft landing, and acrobatic maneuvers such as wall jumps.

\printbibliography

\end{document}